\documentclass[twoside]{article}

\usepackage{amsmath}
\usepackage{amssymb}
\usepackage{chngcntr}
\usepackage{amsthm}
\usepackage{mathtools}
\usepackage{pifont}
\usepackage{dsfont}
\usepackage{bm}
\usepackage{mathrsfs}
\usepackage{eurosym}
\usepackage{tabularx}
\usepackage{booktabs}
\usepackage{cooltooltips}
\usepackage{graphicx}
\usepackage{svg}
\usepackage{colortbl}
\usepackage{color}
\usepackage{tikz}
\usepackage{subfigure}
\usepackage{wrapfig}
\usepackage{comment}
\usepackage{nicefrac}
\usepackage{subcaption}
\usepackage[font=small]{caption}

\usepackage[utf8]{inputenc} 
\usepackage[T1]{fontenc}    
\usepackage{hyperref}       
\usepackage{url}            
\usepackage{booktabs}       
\usepackage{amsfonts}       
\usepackage{nicefrac}       
\usepackage{microtype}      
\usepackage{xcolor}         
\usepackage{algorithm}
\usepackage{algorithmic}
\usepackage[capitalise]{cleveref}
\usepackage{enumitem}
\usepackage{natbib}

\DeclareMathOperator*{\argmax}{arg\,max}
\DeclareMathOperator*{\argmin}{arg\,min}

\newtheorem{theorem}{Theorem}

\newtheorem{proposition}{Proposition}
\newtheorem{corollary}{Corollary}
\newtheorem{definition}{Definition}

\newtheorem{properties}{Properties}
\newtheorem{remark}{Remark}

\newenvironment{derivation}{\emph{Derivation.}\,}{\hfill$\square$}
\usepackage[accepted]{aistats2025}
\begin{document}

%

%

\twocolumn[


\aistatstitle{Additive Model Boosting: New Insights and Path(ologie)s}
\aistatsauthor{ Rickmer Schulte \And David R\"ugamer }

\aistatsaddress{ Department of Statistics, LMU Munich \\
Munich Center for Machine Learning \And  Department of Statistics, LMU Munich \\
Munich Center for Machine Learning } ]

\begin{abstract}
Additive models (AMs) have sparked a lot of interest in machine learning recently, allowing the incorporation of interpretable structures into a wide range of model classes. 
Many commonly used approaches to fit a wide variety of potentially complex additive models build on the idea of boosting additive models.
While boosted additive models (BAMs) work well in practice, certain theoretical aspects are still poorly understood, including general convergence behavior and what optimization problem is being solved when accounting for the implicit regularizing nature of boosting. In this work, we study the solution paths of BAMs and establish connections with other approaches for certain classes of problems. 
Along these lines, we derive novel convergence results for BAMs, which yield crucial insights into the inner workings of the method. 
While our results generally provide reassuring theoretical evidence for the practical use of BAMs, they also uncover some ``pathologies'' of boosting for certain additive model classes concerning their convergence behavior that require caution in practice.
We empirically validate our theoretical findings through several numerical experiments.
\end{abstract}

\section{INTRODUCTION}

Additive models (AMs) are widely used in the statistics and machine learning community. Examples include interpretable boosting methods such as \cite{lou2012intelligible,nori2019interpretml} or the so-called neural additive models \citep[NAMs;][]{agarwal2021neural,radenovic2022neural}. In order to maintain the interpretability of such models in the presence of a high-dimensional feature space, sparsity approaches like the Lasso \citep{tibshirani1996regression} have become indispensable. While methods such as the Lasso are well-understood from a theoretical point of view, their applicability is often restricted to a specific class of problems. An alternative approach that is applicable to a much larger class of AMs and can induce sparse structures even in the presence of complex feature effects is to boost additive models.

\textbf{Boosting Additive Models}\,\,
Gradient boosting methods that use additive model components as base learners have been studied under different names and in different forms in recent years. Under the name of model-based boosting \citep{hothorn2010mboost}, this idea was proposed as an alternative optimization and selection routine for AMs. The resulting boosted additive models that we abbreviate with \emph{BAMs}, allow the fitting of a large plethora of different model classes, including non-linear effects through regression spline representations \citep{schmid2008}, time-varying or functional response models \citep{brockhaus2017boosting}, boosted densities \citep{maier2021additive} or shapes \citep{stocker2023functional}, and have been used for high-dimensional settings both in the context of classical generalized additive models \citep[GAMs;][]{hothorn2006model}, quantile regression \citep{fenske2011identifying} and distributional regression \citep{mayr2012generalized}. Yet, most papers building on the seminal work of \cite{BuehlmannYu2003} and \cite{friedman2001} do not provide theoretical guarantees but instead rely on the corresponding findings in simpler models. 

\textbf{Related Literature}\,\,
Previous work that investigated BAMs from a theoretical perspective make use of the connection to functional gradient descent \citep{friedman2001}, which allows 
providing numerical convergence results of various boosting methods \citep{collins2002, mason1999short, meir2003,  raetsch2001leveraging, ZhangYu2005}. 
Most of these results, however, require rather strong assumptions about the objective function or consider modified versions of boosting by imposing certain restrictions on the step size and considered function space. This limits the applicability of results 
and does not generalize to more complex cases of BAMs. More recent investigations \citep{karimi2016, freund2017, locatello2018} exploit new theoretical insights derived from greedy coordinate descent routines, but are again restricted to specific BAM classes, such as linear models. Accelerated and randomized versions were investigated in \cite{lu2020accelerating, lu2020randomized}. Statistical properties of BAMs that incorporate the iterative fitting in boosting have been investigated in \cite{BuehlmannYu2003, BuehlmannHothorn2007, schmid2008, stankewitz2024early}. However, as we show in our work, these results do not cover important theoretical aspects required for practical applications of BAMs.

\textbf{Problem Statement}\,\, Although empirical results frequently show that more complex BAMs perform well in practice, there is a significant gap between the existing theory for simpler models and the application of these in more complex models. This gap raises the risk of researchers and practitioners using these models without fully understanding what is being optimized and what should be kept in mind when using them in practice.

\textbf{Our Contributions}\,
In this work, we study BAMs from a theoretical perspective and find several connections to other prominent optimization methods. However, our investigations also uncover pathologies inherent to boosting certain additive model classes that have important implications for practical usage.
In summary, our findings include:
\vspace*{-2mm}
\begin{enumerate}
    \item Derivation of exact parameter paths for various $L_2$-Boosting variants.
    \item Explicit characterization of the implicit regularization of several BAM classes, formally showing their difference to explicit regularized counterparts.
    \item Convergence guarantees including a linear convergence rate for (greedy) block-wise boosting.
    \item Specific convergence results for various spline and exponential family boosting models.
\end{enumerate}
\vspace*{-2mm}
We further provide empirical experiments in Section~\ref{sec:experiments} that confirm our theoretical findings.

\vspace{-0.2cm}
\section{Background}\label{sec:background}

\vspace{-0.2cm}

\subsection{Notation}

In this paper, we consider $n$ observations $(y_i,x_i), i\in[n]:=\{1,\ldots,n\}$, that are the realizations of random variables from some joint distribution $\mathbb{P}_{yx}$. We denote the stacked outcome as \mbox{$y = (y_1, \ldots, y_n)^\top \in \mathbb{R}^n$} with $y_i \in \mathcal{Y} \subseteq \mathbb{R}$, and the feature matrix $X = (x_1^\top,\ldots,x_n^\top)^\top \in \mathbb{R}^{n\times p}$ with rows $x_i \in \mathcal{X} \subseteq \mathbb{R}^p$. The $j$th feature in $x_i$ potentially spans multiple columns $p_j$ (e.g., for one-hot encodings) and is denoted with ${x}_{ij} \in \mathbb{R}^{p_j}$, $1\leq p_j \leq p$. Stacking ${x}_{ij}$ for all $n$ observations yields $X_j \in \mathbb{R}^{n\times p_j}$. We will use the notation $\|\cdot\|$ to denote the $L_2$ norm, if not stated otherwise. 
For a matrix $Q$, $\lambda_{max}(Q)$ and $\lambda_{pmin}(Q)$ denote the largest and smallest-non zero eigenvalues of $Q$. 
Our goal is to learn or estimate a parametric model $f(x;\beta)$ that takes features (or predictors) $x$ and given parameters (or weights) $\beta$ produces a prediction. A subset of this parameter is denoted by $\beta_j \in \mathbb{R}^{p_j}$. We measure the goodness-of-fit of $f$ using the loss function $\ell : \mathbb{R}^p \to \mathbb{R}; \; \beta \mapsto \ell(y, f(x, \beta))$. When learning $f$, we will update the model iteratively. In this context, we denote the step size or learning rate with $\nu\in(0,1]$, $\beta^{[k]}$ the value of $\beta$ in the $k$th iteration $k\in\mathbb{N}_0$, and $f^{[k]} = f(x;\beta^{[k]})$. If not stated otherwise, we consider the problem 
\vspace*{-1mm}
\begin{equation}{\label{unconst_prob}}\argmin_{\beta\in\mathbb{R}^{p}} \ \ell(\beta).
\end{equation}
As we investigate boosting of AMs, we choose $f$ to be an additive model as discussed in the following.
\vspace*{-1mm}
\subsection{Additive Models} \label{sec:AMs}
\vspace*{-1mm}
Given a pre-specified loss function, we make the optimization problem in \eqref{unconst_prob} explicit by defining the model class to be a (generalized) additive model of the form
\begin{equation} \label{eq:AM}
    \mathbb{E}(Y|x) = h(f(x; \beta)) = h\left(\textstyle\sum_{j=1}^J f_j(x; \beta_j)\right),
\end{equation}
with $Y$ the random variable related to the observations $y_i$ and $f$ the additive predictor that is additive in the single predictor functions $f_j$ with parameter $\beta_j\in\mathbb{R}^{p_j}$. 
Usually each component $f_j$ encompasses only a subset of features ${x}_{\cdot j}\in \mathbb{R}^{p_j}$. 
For GAMs and BAMs, these functions are typically interpretable by nature (linear effects, univariate splines, low-dimensional interaction terms, etc.). $h$ is a monotonic activation (or inverse-link) function, mapping the learned function values to the domain of $Y$. Examples for $h$ are the sigmoid function for logistic additive models or $h(\cdot) = \exp(\cdot)$ for a Poisson (count) regression model. To ensure interpretability of $f$, we require the $f_j$ functions to be elements of a vector space $\mathcal{F}$, i.e., with closure under addition and scalar multiplication. In other words, if $f_j,g\in\mathcal{F}$ and $c\in\mathbb{R}$, then $f_j + c g \in \mathcal{F}$. 

\textbf{Linearity}\, More specifically, we consider the space of functions $f_j$ that are \textit{linear} in their parameters $\beta_j$. This includes linear feature effects, dummy-encoded binary or categorical variable effects, regression splines such as P-splines \citep{eilers1996}, Kriging \citep{oliver1990kriging}, Markov random fields \citep{rue2005gaussian}, tree-stumps or built trees with leaf weights $\beta_j$, and transfer-learning neural basis functions as in \cite{agarwal2021neural} where $\beta_j$ are the last layer's weights.  For the ease of presentation, we will use 
\begin{equation} \label{eq:simplification}
    f_j(x;\beta_j) = {x}_{\cdot j}^\top \beta_j
\end{equation}
and assume that the aforementioned basis transformations are already encoded in ${x}_{\cdot j}$.

\subsection{Boosting Additive Models}\label{sec:boost_add_models}
\vspace*{-1mm}
In this work, we study a commonly used variant of the gradient boosting algorithm  
proposed in \cite{friedman2001}. 
In \cref{algo:GB}, we present this extension of gradient boosting \citep{BuehlmannHothorn2007}, adapted for the use case of BAMs with so-called base learners $f_j$ as given in \cref{eq:simplification}. While the original algorithm was presented as boosting in function space, the linearity of the base learners $f_j$ in the parameters $\beta_j$
also allows the interpretation of boosting in parameter space.
\begin{algorithm}
\caption{Gradient boosting of additive models} \label{algo:GB}
\small
\begin{algorithmic}[1] 
\STATE Initialize $f^{[0]}$. Set $k=0$.
\STATE Given loss function $\ell(\cdot)$, compute the functional derivative evaluated at the current estimate $f^{[k]}$, i.e.,
\begin{equation}{\label{graboost_s2}}
    \widetilde{y}_{i}^{[k]} := \left. -\frac{\partial}{\partial f} \ell(y_{i}, f) \right|_{ f=f^{[k]}({x}_i)} \: \: i = 1,\ldots,n
    \vspace*{-3mm}
\end{equation}
Set $k = k+1$.
\vspace*{1mm}
\STATE \begin{enumerate}[wide, labelwidth=!, labelindent=0pt]
    \item[a)] For $j=1,\ldots,J$, estimate:
    \vspace*{-3mm}
    \begin{equation*}
        \hat{\beta}_j = \argmin_{\beta_j\in\mathbb{R}^{p_j}} n^{-1} \sum_{i=1}^n (\widetilde{y}_i^{[k]}-{x}_{ij}^\top\beta_j)^2 + \lambda_j \beta_j^\top P_j\beta_j
    \end{equation*}
    \vspace*{-5.5mm}
    \item[b)] Select $\hat{\jmath} = \argmin_{j\in[J]} n^{-1} \textstyle \sum_{i=1}^n (\widetilde{y}_i^{[k]}-{x}_{ij}^\top \hat{\beta}_j)^2$
    \vspace*{1mm}\\
\end{enumerate}
\STATE Update $f^{[k]}(x) = f^{[k-1]}(x) + \nu  \cdot  f_{\hat{\jmath}}(x;\hat{\beta}_j) \,$ with $\nu \in (0,1]$.
\vspace*{-2mm}
\STATE Repeat steps 2 -- 4 until convergence or until a pre-specified stopping criterion is met.
\end{algorithmic}
\end{algorithm}

To use the algorithm as a fitting procedure for models described in Section~\ref{sec:AMs}, one needs to bring the response on the level of the additive predictor $f$. This can be done by defining the loss as $\ell(\beta) := \ell(h^{-1}(y),f(x;\beta))$ for GAMs. More generally, for boosting distributional regression models \citep{mayr2012generalized}, $\ell$ represents the negative log-likelihood of a parametric distribution and the distribution parameters of interest are each modeled via different additive predictors. 
Apart from different base learners and loss functions that can be used in Algorithm~\ref{algo:GB}, fitting can be done \textit{jointly} or in a \textit{greedy block-wise} fashion, thereby resembling a wide variety of gradient boosting variants.

Note, that the original gradient boosting algorithm of \cite{friedman2001} incorporates an additional line search between step 3 and 4 of Algorithm~\ref{algo:GB}. However, this is usually omitted in practice due to its computational costs while only yielding negligible differences in the estimated models, 
especially for small step sizes. 

\subsection{Joint or Block-wise Selection and Updates}\label{sec:joint-block-updates}
The greedy base learner selection of Algorithm~\ref{algo:GB} chooses the best-performing additive component $f_j$ at each step. As a single component $f_j$ may comprise multiple columns of the design matrix $X$, e.g. for splines or categorical variables, we further introduce the notion of block-wise updates with blocks $b\in\mathcal{B} \subseteq [J]$.
Without loss of generality, we assume that each block $b$ corresponds to one set of parameters $\beta_j, j\in[J]$ with corresponding features matrices $X_b = X_j$.  

\textbf{Model Types}\, In step 3a) each base learner is fitted against the current negative gradient $\tilde{y}$. Fitting may vary for each base learner. For penalized base learners, the minimization includes a penalty parameter $\lambda_j >0$ and penalty matrix $P_j \in \mathbb{R}^{p_j\times p_j}$. For unpenalized base learners, we set $\lambda_j=0$. Step b) then chooses the best-fitting base learner (in terms of the $L_2$ loss).
In combination with early stopping, this can enable variable selection as some components $f_j$ might not be selected. In contrast, \textit{joint} updates fit all components simultaneously to the negative gradient such that all components are selected and updated in each step. This can be seen as a special case of the block-wise procedure with a single block, i.e., $J=1$. We theoretically investigate both model types and BAMs with several different base learners and loss functions in \cref{sec:theo_prop_bams}.


\section{THEORETICAL PROPERTIES OF BOOSTED ADDITIVE MODELS}\label{sec:theo_prop_bams}
\begin{figure*}[ht!]
    \vspace*{-3mm}
    \centering
    \includegraphics[width=0.85\textwidth]{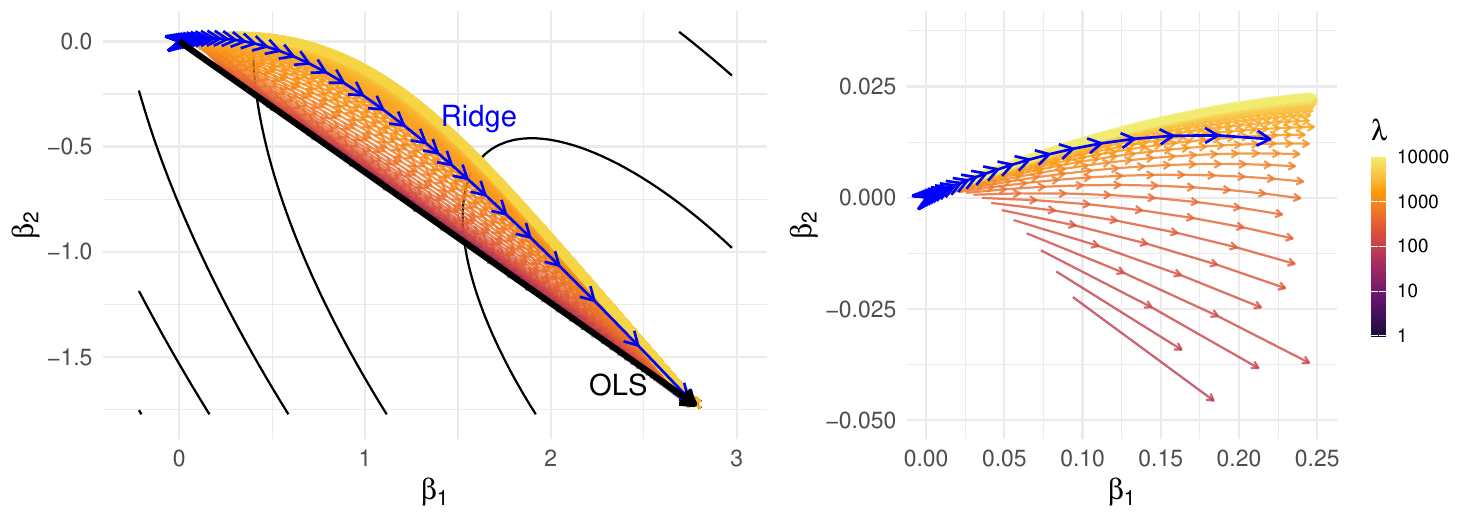}
    \vspace*{-0.3cm}
    \caption{\textit{Left: Paths of boosted linear models with ridge penalty (ridge boosting) for different penalty parameters $\lambda$ (colored lines according to the legend) together with the ridge regression path (blue). Path of linear model boosting is the limiting case of ridge boosting with $\lambda=0$ (black). Block contour lines represent the loss surface. Right: Same plot but zoomed in.}}
\label{fig:ridge_crossing}
\vspace*{-3mm}
\end{figure*}
As with other iterative methods such as the Lasso, the hope is that the boosting algorithm produces an interesting set of submodels among which we can choose the final model. However, whether the obtained paths of parameters $\beta^{[k]}$ are sensible and useful is largely determined by the convergence behavior of the method, which in parts is still poorly understood. Therefore, we investigate whether and under what conditions these paths converge and what solution they are converging to. This also requires studying the implicit regularization of the method and discussing to what extent the method can be related to explicitly regularized optimization problems.



\vspace*{-1mm}
\subsection{Joint Updates}

\subsubsection{Warm Up: Linear Model Boosting}
The simplest case of Algorithm~\ref{algo:GB} is boosting with linear learners, i.e., $f = X\beta$, and $L_2$ loss. Studying this algorithm is instructive for a better understanding of more complex versions.
Boosting with a linear model learner aims to iteratively find the parameters \mbox{$\beta \in \mathbb{R}^{p}$} that minimize the $L_2$ loss.
In combination with joint updates, this is a particular variant of the well-known $L_2$-Boosting \citep{BuehlmannYu2003}. As the negative functional gradient at the current function estimate $\hat{f}^{[k]} = X\beta^{[k]}$ in \eqref{graboost_s2} corresponds to the model residuals, $L_2$-Boosting corresponds to iterative least-squares fitting of the residuals \citep{BuehlmannYu2003}. 
In case $X$ is of full column rank, a simple derivation shows that the parameters in each boosting iteration $k$ can be written as
\vspace*{-0.5mm}
    \begin{align}\label{eq:linear_mod_boost}
        \beta^{[k]} & = \left(\sum_{m=0}^{k-1} \nu (1-\nu)^{m}\right) \cdot \beta^{OLS} :=  \delta^{[k]} \cdot \beta^{OLS}
        \vspace*{-2mm}
    \end{align}
where \mbox{$\nu \in (0,1]$} denotes the step size and \mbox{$\beta^{OLS} := (X^{\top}X)^{-1}X^{\top}y$} the OLS solution (cf.~\cref{app:linear_mod_boost}). The shrinkage is determined by the factor $\delta^{[k]}$, which only depends on $\nu$ and $k$ and is strictly decreasing in both. As the factor can be simplified to $\delta^{[k]} = 1 - (1-\nu)^{k}$, we get convergence to the respective OLS estimator in the limit ($\beta^{[k]} \to \beta^{OLS}$ as $k \to \infty$). This result is to be expected given that linear model boosting with joint updates can be seen to resemble Newton steps (with step size $\nu$) for which convergence on quadratic problems is known \citep{boyd2004}. The convergence of linear model boosting is depicted in Figure~\ref{fig:ridge_crossing}. The same figure also depicts the regularization path of ridge regression and the convergence paths of ridge boosting, a regularized version of linear model boosting. The explicit regularization of ridge regression and boosting's implicit regularization are discussed next.

\begin{figure*}[ht!]
\vspace*{-2mm}
\centering
\includegraphics[width=0.95\linewidth]{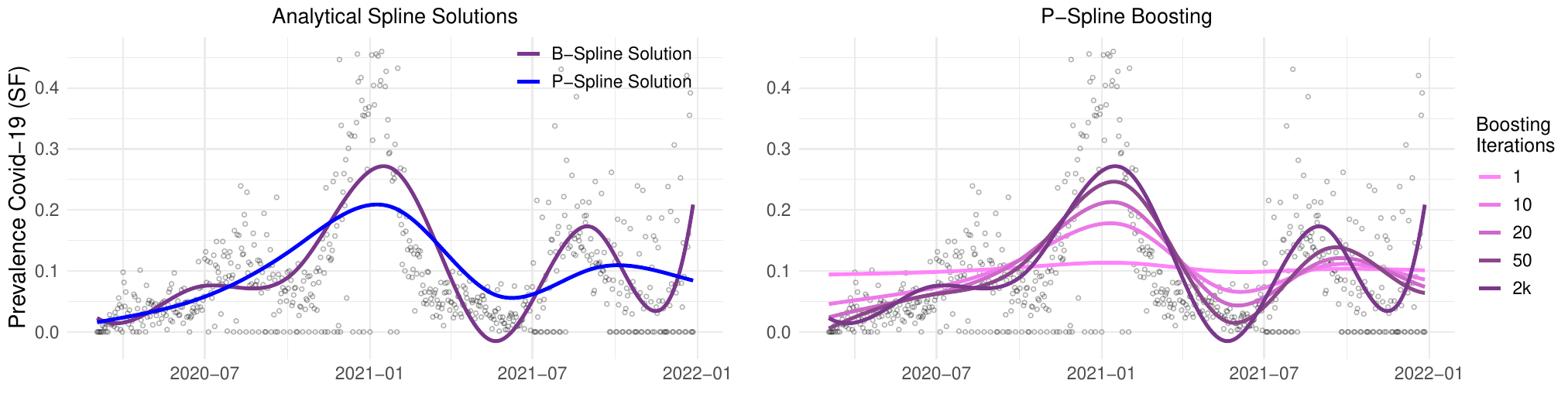}
\vspace*{-6mm}
\caption{\label{fig:p-spline} \textit{Estimated logarithmic Covid-19 prevalence in San Francisco (SF) via BAMs. Left: Analytical B-spline (purple) and P-spline solution (blue). Right: P-spline boosting iterates converge to the unpenalized (B-spline) solution.}}
\vspace*{-3mm}
\end{figure*}

\subsubsection{Boosting with Quadratic Penalties} \label{sec:quadraticpen}

A common alternative to linear models in BAMs are regression splines, which allow for non-linear modeling of features. As an example for regression splines, we use P-splines \citep{eilers1996} which utilize B-splines \citep{deBoor1978} as basis expansion of the predictor variables and penalize (higher-order) differences between adjacent weights $\beta$. For details on P-splines, we refer to Appendix~\ref{app:css_psplines}. Generally, splines can be formulated as an $L_2$ loss optimization problem with quadratic penalization term:
\begin{equation}{\label{p_spline_problem}}
     \min_{\beta \in \mathbb{R}^p}\frac{1}{2} \|y - X\beta\|^{2}+ \frac{\lambda}{2} \beta^{\top}P\beta,
\end{equation}
where $X\in\mathbb{R}^{n\times p}$ is a matrix containing the evaluated basis functions, $\lambda>0$ defines the regularization parameter, and $P \in \mathbb{R}^{p\times p}$ is a symmetric penalty matrix such as a second order difference matrix \citep{Green1994,Wahba1990}. Boosting splines amounts to replacing $y$ with the current negative gradient in \eqref{p_spline_problem} as shown in Algorithm~\ref{algo:GB}.
As regression splines use (penalized) linear effects of multiple basis functions to represent non-linear functions of single features, joint updates (of all basis functions together) naturally arise in their application. Their parameter paths and shrinkage can again be characterized exactly:
\begin{proposition} \label{prop_pspline_joint}
    The estimates of $L_2$-Boosting with quadratic penalty and joint updates in iteration $k$ are given by
    \begin{equation}{\label{eq:shrinkage_pspline}}
        \beta^{[k]}
= [\textstyle\sum_{m=0}^{k-1} \nu (I - \nu (X^{\top}X + \lambda P)^{-1}X^{\top}X)^{m}] \ \beta^{PLS},
    \end{equation}

    with step size $\nu \in (0,1]$, $\lambda>0$, $P$ a symmetric penalty matrix, and the penalized least squares solution \mbox{$\beta^{PLS}:=(X^{\top}X + \lambda P)^{-1}X^{\top}y$}. If $X$ has full column rank,
$$\beta^{[k]} \overset{k \to \infty}{\longrightarrow} (X^{\top}X)^{-1}X^{\top}y =\beta^{OLS},$$
otherwise parameters converge to the min-norm solution $\beta^{[k]} \overset{k \to \infty}{\longrightarrow} X^{+}y$, with $X^{+}$ being the pseudoinverse of $X$.
\end{proposition}

\begin{remark}
    The above result also holds for the special case of ridge boosting with $P = I$. Its parameter paths and convergence to the OLS are shown in Figure \ref{fig:ridge_crossing}.
\end{remark}

The proof of \cref{prop_pspline_joint} builds on the fact that the shrinkage in \eqref{eq:shrinkage_pspline} can be recognized as a Neumann series. Interestingly, despite the explicit regularization, this series converges such that the shrinkage vanishes and the unpenalized fit is obtained in the limit.
Figure~\ref{fig:p-spline} demonstrates the latter using the example of boosting P-splines to model Covid-19 prevalence in San Francisco. 
Details are discussed in \cref{sec:conv-unpen}. 
%

The previous results show that $L_2$-Boosting (with and without penalty)
induces an implicit shrinkage on the parameters. 
The next theorem shows that this implicit shrinkage can be characterized exactly. 
It demonstrates that the boosted parameters also correspond to the unique solution of an explicitly regularized problem. 
\vspace*{1mm}
\begin{theorem}\label{thm:exp_reg_boost}
Given full column rank matrix $X$, $L_2$-Boosting with quadratic penalty and joint updates \eqref{eq:shrinkage_pspline} uniquely solves at each iteration $k \in \mathbb{N}$ the explicitly regularized problem
\begin{equation}\label{eq:exp_reg_boost}
    \min_{\beta \in \mathbb{R}^p} \frac{1}{2} \|y - X\beta\|^{2}+ \frac{1}{2} \beta^{\top}\Gamma_k\beta,
\end{equation}
with $\Gamma_k := (X^{\top}X) \ S_{\lambda}^{-1}  \ [ (I - \nu S_{\lambda})^{-k} - I]^{-1} \ S_{\lambda}$ as penalty matrix and $S_{\lambda} := (X^{\top}X + \lambda P)^{-1}X^{\top}X$.
\end{theorem}
%
\begin{remark}
    The problem in \eqref{eq:exp_reg_boost} changes in each step $k$ and generally does not correspond to the problem in \eqref{p_spline_problem}. Thus, \cref{thm:exp_reg_boost} disproves a widespread interpretation of regularized spline boosting that the algorithm implicitly minimizes \eqref{p_spline_problem} for a specific $\lambda>0$ in each iteration and thereby implicitly finds the model with optimal $\lambda$.
\end{remark}
\cref{thm:exp_reg_boost} also allows us to further investigate the implicit shrinkage of linear model and ridge boosting by determining under what conditions their paths correspond to the solution of ridge regression.
\begin{corollary}\label{matchridge}
     The parameter paths of $L_2$-Boosting for linear models with joint updates correspond to the solutions of ridge regression with penalty parameter
\begin{equation}
    \tilde{\lambda}(k) := \sigma_X^{2} \frac{(1-\nu)^k}{1- (1-\nu)^k}
\end{equation}
if and only if $X^\top X=\sigma_X^{2}I$ for some $\sigma_X^{2}>0$.
\end{corollary}
\begin{remark}
    The same holds for ridge boosting with a slightly different penalty parameter (cf.~\cref{app:matchridge}). Importantly, these equivalences only hold in the case of isotropic features. The fact that paths of boosting and ridge regression differ in general is shown in Figure \ref{fig:ridge_crossing}. Details about Figure~\ref{fig:ridge_crossing} can be found in \cref{app:path}. 
\end{remark}
\subsection{Block-wise Boosting}
We now turn to the block-wise boosting variant and a second main result. As discussed in \cref{sec:joint-block-updates}, the block-wise setting generalizes joint and component-wise updates, including both as special cases. 
Before deriving our main convergence results, 
we first relate this greedy update variant to a specific optimization procedure:
%
\begin{proposition}\label{prop:equi_gbcd}
    Boosting additive models with greedy block-wise updates and $L_2$ loss corresponds to optimizing AMs with greedy block coordinate descent (GBCD) and the Gauss-Southwell-Quadratic (GSQ) update scheme (\ref{block_update_2},\ref{eq:gsq}). In the component-wise case, it matches greedy coordinate descent (GCD) with Gauss-Southwell-Lipschitz (GSL) update scheme (\ref{gsl-rule},\ref{gsl-update}).
\end{proposition}

\begin{remark}\label{rm:equi_gbcd}
    In case of the penalized $L_2$ loss \eqref{p_spline_problem}, block-wise boosting additive models can be seen as G(B)CD with GSQ (GSL) on the unpenalized $L_2$ loss. We give further details in \cref{sec:reg-splines}. The connection in case of other loss functions is discussed in \cref{sec:glmboosting}.
\end{remark}

While GBCD has a long-standing history in the optimization literature, the two mentioned update schemes of GSQ and GSL were only introduced recently and shown to be particularly efficient \citep{nutini2015,nutini2022}. Thus, to the best of our knowledge, the described link between these update schemes and BAMs has not been established so far. Under certain conditions on the loss function, we will use the above equivalence to derive novel convergence guarantees for BAMs with block-wise updates, including a linear convergence rate. For this, we assume the problem to be $\mu$-PL and $L$-smooth in the parameters $\beta \in \mathbb{R}^{p}$ (cf.~\cref{app:funprop} for a definition). Requiring only the weaker condition of $\mu$-PL instead of strong convexity allows us to study convergence also in cases where the optimal solution is not unique, e.g.\ boosting high-dimensional linear models with $n<p$. 

\begin{theorem} \label{thm:block_conv} 
If $\ell(\beta)$ is $\mu$-PL and $L$-smooth with respect to the parameters $\beta$, block-wise boosting with step size $\nu$ converges with rate $\gamma$:
    \vspace*{-2mm}
    \begin{equation} \label{conv_bgcd}
    \ell(\beta^{[k]})-\ell^{*} \leq {\underbrace{\left(1- \nu \frac{\mu}{L_{\mathcal{B}} |\mathcal{B}|} \right)}_{=:\tilde{\gamma}}}^{k}\left(\ell(\beta^{[0]})-\ell^{*}\right),
\end{equation}
for $\nu \in (0,1]$ s.t.\ $\nabla_{bb}^{2} \ell(\beta) \preceq \frac{1}{\nu}X_{b}^{\top}X_{b}$ $\forall \beta \in \mathbb{R}^p$, $\forall b \in \mathcal{B}$.
Above $\ell^{*}$ denotes the optimal loss, 
$|\mathcal{B}|$ the number of blocks, and $L_{\mathcal{B}}$ the largest Lipschitz constant of all blocks with $L_{\mathcal{B}} := \max_{b\in \mathcal{B}} L_{b}\leq L$.
\end{theorem}
\begin{remark}
    A distinct property of boosting additive models, that originates from the quadratic approximation of the negative functional gradient in the boosting algorithm, is the scaling of each gradient update by a particular (block) matrix. 
    Independent of $\ell$, this scaling matrix is the inverse of $\frac{1}{\nu}X_{b}^{\top}X_{b}$ and $\frac{1}{\nu}(X_{b}^{\top}X_{b} + \lambda P_b)$ for unpenalized and penalized base learner, respectively. 
    Hence, for convergence guarantees as in \cref{thm:block_conv}, one requires $\frac{1}{\nu} X_{b}^{\top}X_{b}$ to provide a valid upper bound to the respective block of the Hessian for all $b \in \mathcal{B}$. For some BAM classes, this requires choosing $\nu$ sufficiently small. For others, this upper bound condition is naturally fulfilled. 
    As quadratic problems are $\mu$-PL and $L$-smooth (cf.~\cref{app:funprop}), we get the following corollary.
\end{remark}
\begin{corollary}\label{corr:quad_prob}
    If $\ell(\beta)$ is a quadratic problem with positive semi-definite Hessian $Q$, block-wise boosting converges for $\nu \in (0,1]$:
\vspace*{-1mm}
\begin{equation} \label{conv_bgcd_quad}
    \ell(\beta^{[k]})-\ell^{*} \leq {\underbrace{\left(1- \frac{\nu (2 - \nu)}{|\mathcal{B}|} \frac{\lambda_{pmin}(Q)}{\lambda_{max}(Q)} \right)}_{=:\gamma}}^{k}\!\!\left(\ell(\beta^{[0]})-\ell^{*}\right)\!.
\end{equation}
\end{corollary}
\vspace*{-2mm}
We obtain several insights from \cref{thm:block_conv} and \cref{corr:quad_prob} related to boosting's convergence speed. First, $\gamma$ and $\tilde{\gamma}$ are monotonically decreasing in the step size $\nu$ and increasing in the number of blocks $|\mathcal{B}|$. Given the terms in both convergence rates, it is evident that $\gamma, \tilde{\gamma} \in [0,1)$, so progress is guaranteed in each step. 
Further, the last term of the rate in \eqref{conv_bgcd_quad} is similar to the reciprocal of the condition number of the Hessian $Q$ (using $\lambda_{pmin}$ instead of $\lambda_{min}$). This matches the common notion that gradient methods converge faster for well-posed problems with condition numbers close to one \citep{boyd2004}.

The convergence of component-wise $L_2$-Boosting follows immediately from \eqref{conv_bgcd_quad}, given that the component-wise procedure is just a special case of the block-wise counterpart with $|\mathcal{B}|=p$ and $L_{\mathcal{B}}=L_{CLS}$ where $L_{CLS}$ is the component-wise Lipschitz constant. This result is reassuring as it matches convergence results of \cite{freund2017} which consider the special case of component-wise $L_2$-Boosting with standardized predictors, i.e., $L_{CLS}=1$.


\subsubsection{Regression Splines}\label{sec:reg-splines}

Following \cref{rm:equi_gbcd} (with further discussion in \cref{app:rm_equi_gbcd}), we know that block-wise boosting with penalized loss differs from usual GBCD as it uses the gradient of the unpenalized problem. Similar to the case of joint updates, this means that boosting is neglecting any penalization in previous iterations. As a consequence, the loss in the convergence result of \cref{thm:block_conv} corresponds to the unpenalized loss, leading to convergence to the unpenalized instead of the penalized fit. 
As practitioners are usually interested in the latter, this property might not be desirable. In this case, however, the previous results also provide a way forward by using GBCD instead of boosting as fitting routine of the penalized problem. 
As the result in \cref{thm:block_conv} also holds for the usual GBCD, but in terms of the penalized loss, optimizing penalized regression splines with GBCD will converge to the penalized fit. 


\begin{corollary}\label{corr:conv_regspline}
    Using GBCD with GSQ update scheme to fit regression spline models of the form \eqref{p_spline_problem} that are $\mu$-PL and $L$-smooth in their parameters converges to an optimal solution of the regression spline problem with rate stated in \cref{thm:block_conv}.
\end{corollary}

A prominent alternative fitting procedure for regression splines or additive models is backfitting \citep{friedman1981}. 
While convergence guarantees for backfitting have been analyzed by \cite{buja1989} and \cite{ansley1994}, to the best of our knowledge, no explicit convergence rate has been derived to this date, making the rate in \cref{corr:conv_regspline} the first convergence rate for regression spline problems.

\subsubsection{Cubic Smoothing Splines}\label{sec:css_regspline}

$L_2$-Boosting with cubic smoothing splines (CSS) was proposed in the seminal work of \cite{BuehlmannYu2003}. 
As CSS penalize the second differences of the fitted function, this can be seen as a continuous generalization of boosting with regression splines using second-order difference penalties. 
Using previous findings, we can also establish the 
convergence of component-wise $L_2$-Boosting with multiple CSS (proof in Appendix~\ref{app:multiple_csp}). Interestingly, the result in \cref{prop:css} holds independently of the selection rule (greedy, cyclic, random).

\begin{proposition} \label{prop:css}
  Let $f^{[k]}$ be the fitted function values of component-wise $L_2$-Boosting with cubic smoothing splines after $k$ iterates and $f^\ast$ be the saturated model (perfect fit). Then $f^{[k]} \rightarrow f^\ast$ for $k \rightarrow \infty$.  
\end{proposition}
\begin{figure*}[!h]
\vspace*{-2mm}
\centering
\includegraphics[width=0.75\textwidth]{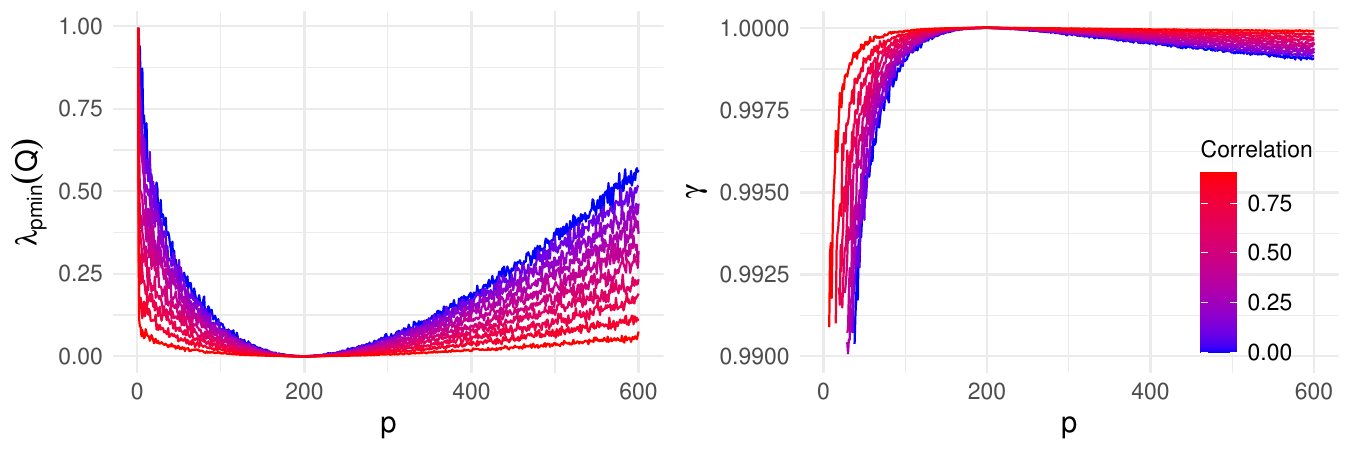}
    \vskip -0.1in
\caption{\textit{Smallest non-zero eigenvalue of $Q$ (left) and convergence rate $\gamma$ as given in \eqref{conv_bgcd_quad} (right) for component-wise boosting for a linear model (hence a quadratic problem) with varying pairwise correlation $\rho$ between predictor variables (color) for fixed $n=200$, $\nu=1$, and varying $p$ (x-axis).}}\label{fig:conv_rate}
\vspace*{-4mm}
\end{figure*}
\subsubsection{Generalized Boosting Approaches} \label{sec:glmboosting}
%
We now show how to extend previous results to models with exponential family distribution assumption. In this case, the density of the response can be written as
\begin{equation}
    p_{\theta}(y) = \exp\left[ \{y \theta - \varsigma(\theta)\}/\phi + c(\phi, y)\right],
\end{equation}
where the terms $\theta$, $\varsigma(\theta)$, $\phi$ and $c(\phi,y)$ depend on the exponential family distribution \citep[see, e.g.,][]{fahrmeir2013, wood2017}. 
The loss function w.r.t.~the functional $f$ is then defined by
\begin{equation}\label{eq:log_lik_expfam}
    \ell(f) = \ell(\theta) = \{y \theta(f) - \varsigma(\theta(f))\}/\phi + c(\phi, y),
\end{equation}
where $f$ is used to model the natural parameter $\theta$. In the so called canonical link case, $f = \theta = h^{-1}(\mathbb{E}(Y|x))$, the log-likelihood in \eqref{eq:log_lik_expfam} can be shown to be strictly convex (cf.~\cref{app:expfam}). For other link functions this property cannot be guaranteed. As strictly convex problems are $\mu$-PL on a compact set \citep{karimi2016}, \eqref{eq:log_lik_expfam} is $\mu$-PL as long as its parameters are bounded. In case the condition of $L$-smoothness is also fulfilled for the respective exponential family problem at hand, e.g., the logistic loss is known to be $\nicefrac{1}{4}$-smooth, then the following proposition holds.

\begin{proposition}\label{prop:expo_conv}
    Block-wise boosting with exponential family loss $\ell$ corresponds to GBCD with GSQ update scheme as long as the Hessian upper bound condition is fulfilled. If $\ell$ is $\mu$-PL and $L$-smooth w.r.t.~the parameters $\beta \in \mathbb{R}^{p}$, the procedure converges with rate as given in \cref{thm:block_conv}.
\end{proposition}

In \cref{sec:exp_glm}, we investigate the examples of Binomial and Poisson boosting. For the latter, we show that a sufficiently small step size is essential to fulfill the upper bound condition and thus to obtain convergence. In contrast, convergence of Binomial boosting follows directly from \cref{prop:expo_conv} for arbitrary step size \mbox{$\nu \in (0,1]$}, as the upper bound condition is naturally fulfilled for the Hessian $Q^{log}$ of the logistic model given that $Q^{log}_{bb} \preceq \frac{1}{4} X_{b}^{\top}X_{b} \preceq X_{b}^{\top}X_{b}$ $\forall b \in \mathcal{B}$.

Next to the derivation of \cref{prop:expo_conv} in \cref{app:expfam_gbcd}, we also discuss other loss functions outside the exponential family class, such as the $L_1$, Huber and Cox proportional hazards loss, in \cref{sec:remarkloss}. 


\subsubsection{Boosting Distributions}\label{sec:dist_boost}
When using BAMs for distributional regression, i.e., learning multiple parameters of a parametric distribution, convexity (and other) assumptions usually do not hold. While a general analysis is challenging, we here study the case of using BAMs to learn both the mean and scale parameters of a Gaussian distribution. To this end, we define the loss function
\begin{equation} \label{eq:loglik_gaussdist}
    \ell(\beta,\xi) = \ell(y,(f_\psi,f_\sigma)) = -\log p_{\mathcal{N}(\psi, \sigma)}(y)
\end{equation} 
as the negative log-likelihood of a Gaussian distribution and parameterize the distribution's mean $\psi = f_\psi(x;\beta)$ and standard deviation $\sigma = f_\sigma(z;\xi)$ both with individual functions $f_\psi(x;\beta) = x^\top \beta$ with parameters $\beta$ and $f_\sigma(z;\xi) = \exp(z^\top \xi)$ with parameters $\xi$. Then the following holds:
\begin{proposition} \label{prop:biconvex}
     The problem \eqref{eq:loglik_gaussdist} is biconvex in $(\beta,\xi)$.
 \end{proposition}
\vspace*{-1mm}
While general convergence guarantees for biconvex optimization are already challenging to obtain \citep{gorski2007biconvex}, another issue that complicates convergence guarantees is that \eqref{eq:loglik_gaussdist} is not $L$-smooth in the parameters $\xi$ (cf.~\cref{app:biconvex}). This can lead to convergence issues as demonstrated in \cref{app:dist_boost_exp}. 

\vspace*{-1mm}
\section{NUMERICAL EXPERIMENTS} \label{sec:experiments}
\vspace{-0.2cm}
In the following, we numerically demonstrate our theoretical findings. 
\vspace{-0.2cm}
\subsection{Convergence Rates}
\begin{figure*}[ht!]
\vspace*{-2mm}
\centering
\includegraphics[width=1\textwidth]{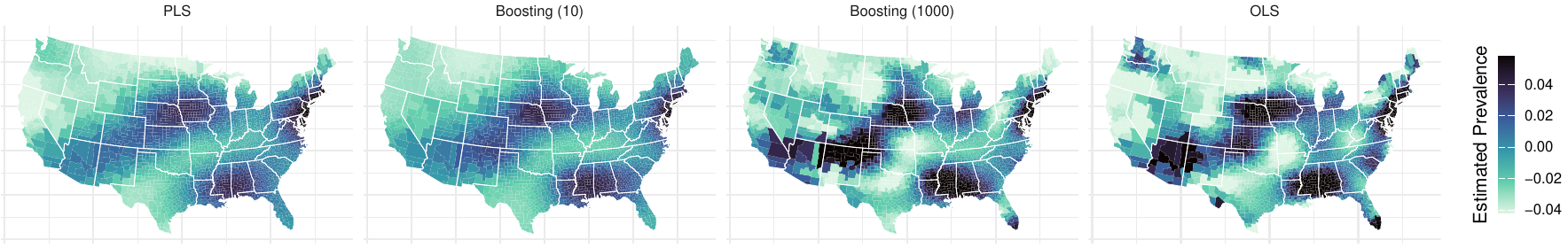}
    \vskip -0.1in
\caption{\textit{Mean-centered spatial effect of Covid-19 prevalence in the United States obtained with BAMs. From left to right: Penalized least squares (PLS); early-stopped BAM; BAM with a large number of iterations; unpenalized least squares fit (OLS) to which boosting is converging to.}}\label{fig:spatial}
\vspace*{-4mm}
\end{figure*}
\vspace*{-1mm}
In Figure~\ref{fig:conv_rate}, we investigate the influence of different degrees of pairwise correlations between predictors on the convergence rate $\gamma$ and the $\mu$-PL constant ($\lambda_{pmin}(Q)$) for a linear model with an increasing number of predictors. 
Generally, $\gamma$ is close to one, reflecting slow convergence, consistent with the typical slow \mbox{(over-)}
fitting behavior of boosting \citep{BuehlmannHothorn2007}.
Lower pairwise correlations lead to faster convergence, but as the number of predictors increases, $\gamma$ rises until $p=n$. In the underdetermined case ($n<p$), $\gamma$ decreases slightly while staying near one, driven by changes in the smallest non-zero eigenvalue of the system.
We further investigate the impact of different condition numbers 
on the linear convergence rate in Figure~\ref{fig:conv_condnum} (in Appendix~\ref{app:further}). In line with the theoretical results obtained from \cref{thm:block_conv}, the convergence is slower for higher condition numbers of the problem.
\vspace*{-2mm}
\subsection{Convergence to the Unpenalized Model}{\label{sec:conv-unpen}}
\vspace*{-1mm}
Another implication of our analysis is the solution the boosting algorithm is converging to. As discussed in previous sections, this corresponds to the unpenalized solution for linear, ridge, (cf.~Figure~\ref{fig:ridge_crossing}) and regression spline boosting (cf.~Figure~\ref{fig:p-spline}). This can have drastic effects in cases where penalized base learners are inevitable such as in spatial modeling. To demonstrate this, we again model the Covid-19 prevalence using BAMs, but now spatially over the entire US.
Figure~\ref{fig:spatial} shows the estimated spatial effect surface of the penalized fit (PLS), boosting after 10 and 1000 iterations, and the unpenalized fit (OLS). Results clearly indicate that boosting can roughly match the PLS estimation, but when running the algorithm further it will --- despite explicit penalization --- converge to the unpenalized model fit. In \cref{app:exper_conv_unpen}, we demonstrate this phenomenon along with various other boosting applications such as boosting with ridge penalty and P-splines as well as the more complex setup of function-on-function regression boosting.
\begin{figure}[!h]
    \centering
    \includegraphics[width=0.45\textwidth]{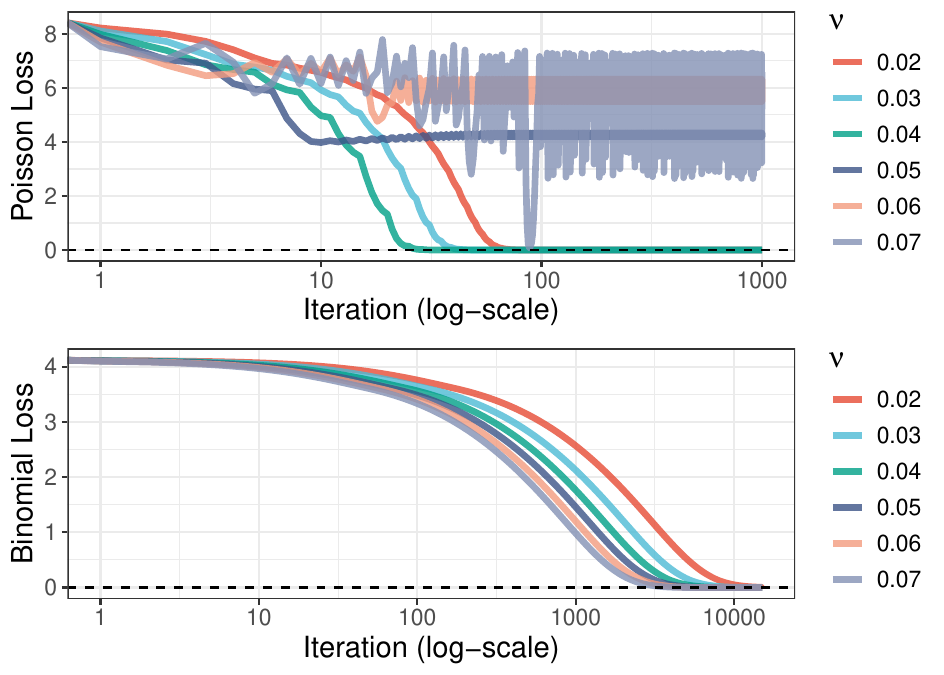}
    \vspace*{-3mm}
    \caption{\textit{Loss path for Poisson (top) and Binomial (bottom) BAMs with different learning rates (colors) showing potential convergence issues for Poisson BAMs.
    }}\label{fig:convergence_pois_binom}
\end{figure}
\subsection{Exponential Family Boosting}\label{sec:exp_glm}
\vspace*{-1mm}
%
%
Lastly, we demonstrate findings from Section~\ref{sec:glmboosting} to show differences in convergence for different distribution families. To this end, we simulate a Poisson and Binomial GLM model and then run different BAMs with learning rates $\nu\in\{0.02,0.03,\ldots,0.07\}$ for both distributions for a maximum of 1000 iterations. 

\textbf{Results:}\, As derived in \cref{prop:expo_conv}, we observe convergence for Binomial BAMs for all learning rates in Figure~\ref{fig:convergence_pois_binom}, whereas for Poisson BAMs, half of the defined learning rates do not upper bound the Hessian correctly and result in oscillating parameter updates and non-convergence. We show the issue of non-convergence for Poisson BAMs also on observational data in \cref{app:glm_exp} and provide further details about the experiments in this and previous sections in \cref{app:further}.
\section{DISCUSSION} \label{sec:discussion}
Boosted additive models (BAMs) are an indispensable toolbox in many applications. Understanding the inner workings of BAMs and their induced implicit regularization is key to insights into their theoretical properties. In this work, we characterize the implicit shrinkage of BAMs by relating them to solution paths of explicitly regularized problems.
We further establish an important link between greedy block-wise boosting and greedy block coordinate descent with a particular update scheme, namely the GSQ update scheme \citep{nutini2015}. Using this equivalence, we derive novel convergence results for BAMs. 


Our investigation also uncovers several pathologies of BAMs. We show that boosting penalized models neglects the penalization in previous steps and therefore converges to the unpenalized fit. For exponential family loss, we show that the implicit Hessian approximation in BAMs might induce non-convergence.

\textbf{Practical Implications}\, Several aspects can be derived from our results, in particular, the effects of penalization, the effects of Hessian approximation and the convergence rates of the optimization routine. While the results on penalization could potentially be used in practice to design a routine with a desired amount of regularization, the effects of Hessian approximation suggest a careful consideration of the choice of learning rate. Although a small step size may be detrimental to the speed of convergence, it is essential to ensure convergence for certain classes of BAMs. This was found to be particularly important for more complex models, e.g. when boosting distributions. 

\textbf{Future Research}\, Furthermore, because our results provide an explicit form of the objective that is implicitly optimized when using boosting for model estimation, it is theoretically possible to use our results to derive statistical inference (i.e., tests and confidence intervals) analogous to inference in Ridge penalty-type models. As BAMs are usually fitted in combination with early stopping \citep{BuehlmannYu2003, ZhangYu2005, BuehlmannHothorn2007}, derived inference statements would also have to account for the additional regularization induced by stopping the boosting procedure before convergence. 

\bibliography{bibliography}


\appendix

\clearpage

\onecolumn

\section*{SUPPLEMENTARY MATERIAL}

\section{PRELIMINARIES}


\subsection{Function Properties} \label{app:funprop}

To derive theoretical convergence results for gradient optimization methods, 
conditions such as strong convexity and ${L}$-smoothness have been considered.
A common notion of smoothness used in convergence analysis \citep[e.g.,][]{boyd2004, nesterov2012, nesterov2004} is the following:
\begin{definition}
\label{def:Lip} A function $\ell: \mathbb{R}^{p} \to \mathbb{R}$ is ${L}$-smooth with ${L} > 0$, if $\forall \beta, \tilde{\beta} \in \mathbb{R}^{d}$:
$$\|\nabla \ell(\beta) - \nabla \ell(\tilde{\beta})\| \leq L \|\beta - \tilde{\beta}\|.$$
\end{definition}
${L}$-smoothness is sometimes also referred to as ${L}$-Lipschitz continuous gradient $\nabla \ell$. From \cref{def:Lip} we can deduce that an ${L}$-smooth function $\ell$ fulfills $\forall \beta, \tilde{\beta} \in \mathbb{R}^{p}$
\begin{equation}{\label{lipschitz_cond}}
    \ell(\tilde{\beta}) \leq \ell(\beta) + \langle \nabla \ell(\beta), \tilde{\beta} - \beta \rangle + \frac{L}{2} \|\tilde{\beta}-\beta\|^{2}.
\end{equation}
By contrast, a function that is strongly convex, or more precisely $\mu$-strongly convex with $\mu > 0$, fulfills $\forall \beta, \tilde{\beta} \in \mathbb{R}^{p}$
\begin{equation}{\label{strong-cvx}}
    \ell(\tilde{\beta}) \geq \ell(\beta) + \langle \nabla \ell(\beta), \tilde{\beta} - \beta \rangle + \frac{\mu}{2} \|\tilde{\beta}-\beta\|^{2}.
\end{equation}
For twice-differentiable objective functions, the conditions in \eqref{lipschitz_cond} and \eqref{strong-cvx} provide lower and upper bounds on eigenvalues of the Hessian, $\mu I \preceq \nabla^{2}\ell(\beta) \preceq  L I \, \forall \beta \in \mathbb{R}^{p}$, 
where $I$ is the identity matrix.
More recently, \cite{karimi2016} showed that it is sufficient to consider the PL-inequality instead of strong convexity to derive convergence rates for iterative gradient methods. 
\begin{definition}{\cite{karimi2016}}
\label{def:PL} A function $f: \mathbb{R}^{p} \to \mathbb{R}$ is ${\mu}$-PL with some ${\mu} > 0$, if $\forall \beta \in \mathbb{R}^{p}$:
\begin{equation}{\label{pl_ineq}}
    \frac{1}{2} \| \nabla \ell(\beta) \|^{2} \: \geq \: \mu \: (\ell(\beta) - \ell^{*}).
\end{equation}
\end{definition}
In \cref{def:PL}, $\ell^{*}$ denotes the optimal function value of the optimization problem in \eqref{unconst_prob}. The notation $\ell^{*}$ instead of $\ell(\beta^{*})$ is used as the optimal function value is unique, whereas the solution $\beta^{*}$ to the problem $\ell(\beta)$ with $\ell^{*} = \ell(\beta^{*})$ does not have to be unique in case $\ell(\beta)$ is $\mu$-PL. This is in contrast to strong convexity, where the solution $\beta^{*}$ is guaranteed to be unique. 
\cref{def:PL} is more general than strong-convexity 
and several important problems do not fulfill strong convexity but the more general PL-inequality. 

\paragraph{Convex Quadratic Problems} The above conditions can be shown to hold for the important class of convex quadratic problems, that can be written in the following form 
\begin{equation}\label{quad_prob}
    \min_{\beta \in \mathbb{R}^{p}} \ell(\beta) = \min_{\beta \in \mathbb{R}^{p}} \frac{1}{2} \beta^{\top}Q\beta + q_{1}^{\top}\beta + q_{0},
\end{equation}
where $Q$ denotes a symmetric positive semi-definite (p.s.d.) matrix. A simple derivation shows that the least squares and P-spline problem can be written in this form with $Q$ corresponding to the Hessian ($Q_{LS} = X^{\top}X$ and $Q_{PLS} = X^{\top}X + \lambda P$). The problem in \eqref{quad_prob} is ${L}$-smooth with some Lipschitz constant
$L \leq \lambda_{max}(Q)$, 
where $\lambda_{max}(Q)$ denotes the largest eigenvalue of $Q$.

Further, \cite{freund2017,karimi2016} showed that any problem that can be written in the form of \eqref{quad_prob} is $\mu$-PL with $\mu = \lambda_{pmin}(Q)$, where $\lambda_{pmin}(Q)$ denotes the smallest non-zero eigenvalue of $Q$. 
For p.d.\ $Q$ (all eigenvalues positive), we can even recover the stronger condition of strong convexity, as we get that $\mu = \lambda_{min}(Q) > 0$. This indeed makes a difference e.g. when considering the least squares problem in the overdetermined $n > d$-setting ($X^{\top}X$ p.d.) compared to the underdetermined setting with $n < d$ ($X^{\top}X$ only p.s.d.) as frequently encountered in high-dimensional statistics.

\textbf{A different formulation of $L$-smoothness} As we are dealing with matrix updates and not just single coordinate updates in our derivations, it will prove beneficial to state the ${L}$-smoothness condition in the form of
\begin{equation}{\label{block-lip}}
    \|\nabla_{b} \ell(\beta + U_{b}\upsilon) - \nabla_{b} \ell(\beta)\|_{H_{b}^{-1}} \leq \|\upsilon\|_{H_{b}},
\end{equation}
where $\beta \in \mathbb{R}^{p}$, $H_{b} \in \mathbb{R}^{|b|\times|b|}$ and $\upsilon \in \mathbb{R}^{|b|}$. 
When considering twice-differentiable functions for $\ell$,
\eqref{block-lip} essentially states that $H_{b}$ must provide an upper bound with respect to the block of the Hessian belonging to the coordinates of block $b$, i.e., $\nabla_{bb}^{2} \ell(\beta) \preceq H_{b}$.  
Note that \eqref{block-lip} contains the usual ${L}$-smoothness condition as described in \eqref{lipschitz_cond} as a special case by choosing $H_{b} = L_{b}I$, with $L_{b}$ being the Lipschitz constant for block $b$ \citep{nutini2022}. Conversely, \eqref{block-lip} can be shown to hold, whenever the $L$-smoothness condition \eqref{lipschitz_cond} is assumed to hold, as we have that $L_{b}<L$, $\forall b \in \mathcal{B}$.

\subsection{Optimization Routines}\label{sec_compblock}
In the subsequent convergence analyses, we will consider both component-wise as well as block-wise gradient methods. 

\subsubsection{Greedy (Block) Coordinate Descent}\label{app:gbcd}
Greedy coordinate descent (GCD) with constant step size $\nu \in (0,1]$ is an iterative method in which the update steps are performed component-wise as
\begin{equation}{\label{comp_update}}
    \beta^{[k+1]}=\beta^{[k]}- \nu \nabla_{i_{k}} \ell\left(\beta^{[k]} \right) e_{i_{k}},
\end{equation}
where $e_{i_{k}}$ is the unit vector corresponding to the variable $i_k \in \{1,\ldots,p\}$ selected to be updated at step $k$. 

Greedy block coordinate descent (GBCD) works similarly to GCD but considers blocks of variables instead of single variables to be updated in each step. Thus, the $d$ variables are partitioned into disjoint blocks, where each block is indexed by $b \in \mathcal{B}$. Overall, one obtains a total number of $|\mathcal{B}|$ blocks, where $|\cdot|$ denotes the cardinality. 
Note that by considering a single coordinate per block, we recover GCD, which is why the latter can be seen as a special instance of GBCD. The block-wise updates in GBCD are of the form
%
    $\beta^{[k+1]}=\beta^{[k]} + \nu U_{b_{k}} \varkappa_{b_{k}}$, 
%
with step size $\nu \in (0, 1]$ and 
$U_{b_{k}}$ a block-wise matrix with an identity matrix block for the selected block $b_{k}$ at the current step $k$ and else zeros. With $\varkappa_{b_{k}}$ we denote the direction in which the selected block will be updated. This direction can be chosen to correspond to the block-wise steepest descent direction and thus to the negative gradient with respect to the variables of the selected block 
    $\varkappa_{b_{k}} = - \nabla_{b_{k}} \ell\left(\beta^{[k]} \right)$, 
or can be extended to a matrix update by scaling with matrix $H_{b_{k}}$ to yield
\begin{equation}{\label{block_update_2}}
    \varkappa_{b_{k}} = - (H_{b_{k}})^{-1} \nabla_{b_{k}} \ell\left(\beta^{[k]} \right),
\end{equation}
where $H_{b_{k}}$ could correspond to the respective block of the Hessian or an upper bound of the latter \citep{nutini2022}. 

\subsubsection{Update Rules} \label{app:updaterules}

\paragraph{Gauss-Southwell(-Lipschitz)}

A common selection strategy for the selection of the $i_k$th coordinate in GCD is to use the Gauss-Southwell (GS) selection rule
\begin{equation}{\label{gs-rule}}
    i_{k} = \argmax_{i} |\nabla_{i} \ell(\beta^{[k]})|.
\end{equation}
A more elaborated update routine, the so called Gauss-Southwell-Lipschitz (GSL) rule, was proposed by \cite{nutini2015} and is given by
\begin{equation}{\label{gsl-rule}}
    i_{k} = \argmax_{i} \frac{|\nabla_{i} \ell(\beta^{[k]})|}{\sqrt{L_{i}}}.
\end{equation}
The GSL rule not only takes the gradient into account but also the curvature along each component by scaling with respect to the Lipschitz constant $L_{i}$ of the $i$-th component. This second-order information in the GSL rule can be incorporated into the update step \eqref{comp_update} as well, yielding 
\begin{equation}{\label{gsl-update}}
    \beta^{[k+1]}=\beta^{[k]}- \nu \frac{1}{L_{i_{k}}} \nabla_{i_{k}} \ell\left(\beta^{[k]} \right) e_{i_{k}}.
\end{equation}

\paragraph{Gauss-Southwell-Quadratic}

Analogously to the block update in \eqref{block_update_2}, a greedy block selection rule called Gauss-Southwell-Quadratic (GSQ) is defined by
\begin{equation} \label{eq:gsq}
\begin{aligned}
    b_{k} \quad = \quad & \argmax_{b \in \mathcal{B}} \quad \{\|\nabla_{b} \ell(\beta^{[k]})\|_{{H_{b}}^{-1}}\}, 
\end{aligned}
\end{equation}
where, $\|\cdot\|_{H} = \sqrt{\langle H\cdot,\cdot\rangle}$ denotes a general quadratic norm (a proper norm as long as $H$ is a positive definite matrix) \citep{nutini2022}.

\subsection{Linear Operators}

We will prove \cref{prop:css} below using operator norms. For this, we will consider linear operators.
\begin{properties} \label{proper:op}
Let $T:\mathbb{R}^{n} \to \mathbb{R}^{n}$ be a linear operator and $z, c \in \mathbb{R}^{n}$. Further let $T_{1}:\mathbb{R}^{n} \to \mathbb{R}^{n}$ and $T_{2}:\mathbb{R}^{n} \to \mathbb{R}^{n}$ be two operators. Then the following holds:
\begin{enumerate}
    \item{\label{pf:css_pre1}} Operator Norm: $\|T\|^{*}=\underset{\|z\|=1}{\sup}\|T z\| = \underset{\|z\|\neq0}{\sup}\|T\frac{z}{\|z\|}\|$; 
    \item{\label{pf:css_pre2}} $\|T\|\|c\|= \|c\|\|T\frac{c}{\|c\|}\|
    \leq
    \|c\|\underset{\|c\|\neq0}{\sup}\|T\frac{c}{\|c\|}\|=
    \|c\|\|T\|^{*}$;
    \item{\label{pf:css_pre3}} $\|T_{1}T_{2}\|^{*}= \underset{\|z\|\neq0}{\sup}\frac{\|T_{1}T_{2} z\|}{\|z\|} = \underset{\|z\|\neq0}{\sup}(\frac{\|T_{1}T_{2} z\|}{\|T_{2}z\|}\frac{\|T_{2} z\|}{\|z\|})\leq \underset{\|T_{2}z\|\neq0}{\sup}\frac{\|T_{1}T_{2} z\|}{\|T_{2}z\|}
    \underset{\|z\|\neq0}{\sup}\frac{\|T_{2} z\|}{\|z\|} =
    \|T_{1}\|^{*}\|T_{2}\|^{*}$.
\end{enumerate}
\end{properties}

\subsection{Exponential Family}\label{app:expfam}

For any exponential family model, the density of the response can be written in the following form
\begin{equation}
    p_{\theta}(y) = \exp\left[ \{y \theta - \varsigma(\theta)\}/\phi + c(\phi, y)\right],
\end{equation}
where the terms $\theta$, $\varsigma(\theta)$, $\phi$ and $c(\phi,y)$ depend on the exponential family distribution considered \citep{fahrmeir2013, wood2017}. 
The parameter $\theta$ is called the canonical parameter and is often written as $\theta(\psi)$, due to its dependence on the conditional expectation of the outcome given features, $\mathbb{E}(Y|x)=\psi$. In this setup, one aims to estimate a function $f$, given some response function $h(\cdot)$, such that $h(f_{i})=\mathbb{E}[Y_{i}]=\psi_{i}$. For generalized linear models (GLMs) $f$ is linear in the estimated parameter $\beta$, as we assume $f=X\beta$. For each exponential family, there exists a unique canonical link function $g = h^{-1}$, such that $\theta_{i}=f_{i}$ \citep{fahrmeir2013}. Choosing the canonical link has several theoretical benefits. First, the log-likelihood $\log(p_{\theta}(y))$ used to estimate the model, can be written both in terms of $\theta$ or $f$. Thus, the loss function is defined by
\begin{equation}\label{eq:log_lik_expfam2}
    \ell(f) = \ell(\theta(f)) = \{y \theta(f) - \varsigma(\theta(f))\}/\phi + c(\phi, y).
\end{equation}
Another key feature of the canonical link is that \eqref{eq:log_lik_expfam2} is strictly convex in $f$. For other link functions this property cannot be guaranteed. We can make this notion more explicit by looking at the Hessian of \eqref{eq:log_lik_expfam2}. We do so by considering GLMs, for which \eqref{eq:log_lik_expfam2} becomes a function of the parameter $\beta$. Using the canonical link, the Hessian of the log-likelihood simplifies and coincides with the Fisher information matrix \citep{wood2017}. The latter \citep[derived, e.g., in][]{fahrmeir2013} corresponds to
\begin{equation} \label{eq:hessian_expfam}
    \nabla^{2} \ell(\beta) = X^{\top}WX
\end{equation}
with $W=\text{diag}(\ldots, \tilde{w_{i}},\ldots)$ and 
\begin{equation}{\label{eq:hessian}}
    \tilde{w_{i}} = \frac{{(h'(f_{i}))}^{2}}{\varsigma''(\theta_{i})\phi}.
\end{equation}
By looking at the definition of the respective terms for different exponential family distributions, it becomes clear that $\varsigma''(\theta) > 0$ and $\phi>0$ \citep{fahrmeir2013, wood2017}. With the canonical response function being strictly monotonic, the numerator must be greater than zero as well. Therefore, given the canonical link, the weights are positive and the Hessian positive definite. Thus, as long as $X$ is full rank, the log-likelihood is strictly convex. 

\section{PROOFS AND DERIVATIONS}

\subsection{Derivation of Equation \eqref{eq:linear_mod_boost}}\label{app:linear_mod_boost}

\begin{derivation}
As $L_2$-Boosting for linear models with joint updates corresponds to least squares fitting of residuals from the previous iteration, we can write the parameters at each iteration recursively (with $ \beta^{[0]} := 0$):
\begin{align*}
\beta^{[1]} &= \beta^{[0]} + \nu (X^{\top}X)^{-1} X^{\top}y = \nu \beta^{OLS} \\ 
\beta^{[2]} &= \beta^{[1]} + \nu (X^{\top}X)^{-1} X^{\top} (y - X\beta^{[1]}) = (1- \nu) \beta^{[1]} + \nu \beta^{OLS} = [(1 - \nu) \nu + \nu]\ \beta^{OLS}
\\
\beta^{[3]} &= \beta^{[2]} + \nu (X^{\top}X)^{-1} X^{\top} (y - X\beta^{[2]}) = [(1 - \nu)^{2} \nu + (1 - \nu) \nu + \nu] \ \beta^{OLS}
\\
& \ \ \vdots\\
\beta^{[k]} &= (1 - \nu) \beta^{[k-1]} + \nu \beta^{OLS}\\
&= [\sum_{m=0}^{k-1} \nu (1-\nu)^{m}] \ \beta^{OLS} = \nu \left( \frac{1 - (1-\nu)^{k}}{\nu} \right) \beta^{OLS} = \underbrace{\left(1 - (1-\nu)^{k} \right)\ }_{:= \delta(\nu)^{[k]}}\beta^{OLS},
\end{align*}
where the form of the parameter at iteration $k$ follows by induction. The last equality follows from the fact that we have a geometric series, which converges as $(1-\nu)<1$ with a learning rate $\nu \in (0,1)$. Using this notation, the fitted function at each iteration $k$ can be seen as a linear smoother, by writing $f^{[k]}=\delta(\nu)^{[k]}X(X^{\top}X)^{-1} X^{\top}y$. For $\nu = 1$ we get convergence in one step. Moreover, we can observe that $\delta(\nu)^{[k]} \overset{k \to \infty}{\to} 1$ for arbitrary $\nu \in (0,1)$, such that boosting paths converge to the respective ordinary least squares solution (OLS).
\end{derivation}

\subsection{Proof of \cref{prop_pspline_joint}}

\begin{proof}
As $L_2$-Boosting for linear models with joint updates and quadratic penalization corresponds to penalized least squares fitting of residuals from the previous iteration, we can write the parameters at each iteration recursively (with $\beta^{[0]} := 0$):
\begin{align*}
\beta^{[1]} &= \beta^{[0]} + \nu (X^{\top}X + \lambda P)^{-1} X^{\top}y = \nu \beta^{PLS} \\ 
\beta^{[2]} &= \beta^{[1]} + \nu (X^{\top}X + \lambda P)^{-1} X^{\top} (y - X\beta^{[1]}) = (I - \nu (X^{\top}X + \lambda P)^{-1} X^{\top}X) \nu \beta^{PLS} + \nu \beta^{PLS}
\\
\beta^{[3]} &= \beta^{[2]} + \nu (X^{\top}X + \lambda P)^{-1} X^{\top} (y - X\beta^{[2]}) \\
&= \left[(I - \nu (X^{\top}X + \lambda P)^{-1} X^{\top}X)^{2}\nu + (I - \nu (X^{\top}X + \lambda P)^{-1} X^{\top}X)\nu + \nu\right] \ \beta^{PLS} 
\\
& \ \ \vdots\\
\beta^{[k]} &= (I - \nu (X^{\top}X + \lambda P)^{-1} X^{\top}X) \beta^{[k-1]} + \nu \beta^{PLS}\\
&= [\sum_{m=0}^{k-1} \nu (I - \nu (X^{\top}X + \lambda P)^{-1}X^{\top}X)^{m}] \ \beta^{PLS}
\end{align*}
    where the form of the parameter at iteration $k$ follows by induction. If we let the number of iterations grow to infinity, we get
    $$ \beta^{[k]} \overset{k \to \infty}{\longrightarrow} [\sum_{m=0}^{\infty} \nu (I - \nu (X^{\top}X + \lambda P)^{-1}X^{\top}X)^{m}] \beta^{PLS} = [\nu \sum_{m=0}^{\infty} T^{m}] \beta^{PLS},$$
    where the latter can be recognized as a Neumann series with operator $T:=(I - \nu (X^{\top}X + \lambda P)^{-1}X^{\top}X)$ with $\nu \in (0,1]$. In the following, we differentiate between two cases. First, we discuss the case of $X$ having full column rank ($X^\top X$ is p.d.) and subsequently the rank deficient case where $X$ has reduced column rank ($X^\top X$ is p.s.d.).
    
    \textbf{Full column rank case:} The Neumann series is known to converge if $\|T\|^{\ast}<1$, where $\|\cdot\|^{\ast}$ denotes the operator norm. To show that this is the case, we can first write
    $$\|T\|^{\ast} = \|I - \nu R\|^{\ast} = 1 - \nu \lambda_{min}(R),$$
    with $R:= (X^{\top}X + \lambda P)^{-1}X^{\top}X$. The eigenvalues of $R$ are the same as the eigenvalues of a hat matrix of a penalized regression spline, which are known to be bounded by zero and one \citep{schmid2008}. Further, as $R$ is the product of two symmetric p.d. matrices (as $X^{\top}X$ was assumed to be p.d.), $R$ is positive definite and thus $\lambda_{min}(R)>0$. From this it follows that $\|T\|^{\ast}<1$ and the Neumann series above converges as
    $$ \nu \sum_{m=0}^{\infty} T^{m} = \nu (I - T)^{-1} = \nu ( I - (I - \nu (X^{\top}X + \lambda P)^{-1}X^{\top}X))^{-1} = (X^{\top}X)^{-1} (X^{\top}X+ \lambda P)$$
    Therefore:
    \begin{align*}
        \beta^{[k]} \overset{k \to \infty}{\longrightarrow} [\nu \sum_{m=0}^{\infty} T^{m}] \beta^{PLS} & = (X^{\top}X)^{-1} (X^{\top}X + \lambda P) (X^{\top}X + \lambda P)^{-1} X^{\top}y = (X^{\top}X)^{-1}X^{\top}y
    \end{align*}
\textbf{Rank-deficient case:} Now consider the case that $X$ has reduced column rank. This is a setting that often occurs when a P-spline basis expansion is used with multiple basis functions (determined by the number of knots). In the reduced rank case with $\text{rank}(X)=r<p$, the convergence of the Neumann series can no longer be guaranteed by the same arguments as above, as $\|T\|^{\ast}<1$ may no longer hold. The largest eigenvalue might be one, due to the rank-deficiency. In the following, we consider $P=I$ (ridge boosting) w.l.o.g.\ as any p-spline can be rewritten in terms of a ridge penalty using a modified design matrix. To show convergence in this case, we first pre-multiply $(I-T)$ to the Neumann series above
$$ (I-T) \nu \sum_{m=0}^{\infty} T^{m} = \nu \sum_{m=0}^{\infty} (I-T) T^{m} = \nu \sum_{m=0}^{\infty} T^{m} - T^{m+1}= \nu \left(I - \lim_{m \to \infty} T^{m}\right).$$
The last term can be shown to converge. Using a the full SVD $X=U\Sigma^{\frac{1}{2}} V^\top$, we can write
\begin{align*}
    I - \lim_{m \to \infty} T^{m} &= I - \lim_{m \to \infty} \left(I - \nu  (X^{\top}X + \lambda P)^{-1}X^{\top}X\right)^{m}\\
    &= I - \lim_{m \to \infty} \left(I - \nu  V(\Sigma + \lambda I)^{-1}\Sigma V^\top\right)^{m}\\
    &= I - V\lim_{m \to \infty} \left(I - \nu (\Sigma + \lambda I)^{-1}\Sigma \right)^{m}V^\top\\
    &= I - V D_s^\perp V^\top = V D_s V^\top\\
\end{align*}
with a diagonal matrix $D_s=\text{diag}(1_r^\top, 0_{p-r}^\top)$ containing $r$ ones and $p-r$ zeros on the diagonal, and $D_s^\perp = I - D_s$. Given the previous display, the Neumann series above converges
$$\nu \sum_{m=0}^{\infty} T^{m} = \nu (I-T)^+ V D_s V^\top = (X^\top X)^+ (X^{\top}X + \lambda P) \cdot V D_s V^\top,$$
where $(X^{\top}X)^+$ denotes the Moore–Penrose generalized inverse of $X^{\top}X$ and where we used basic properties of the Moore–Penrose generalized inverse to simplify the expression. Therefore, in the case of rank deficiency, we get convergence of the parameters to the min-norm solution
    \begin{align*}
        \beta^{[k]} \overset{k \to \infty}{\longrightarrow} [\nu \sum\nolimits_{m=0}^{\infty} T^{m}] \cdot\beta^{PLS} 
        & = (X^{\top}X)^{+} (X^{\top}X + \lambda P) \cdot V D_s V^\top \cdot (X^{\top}X + \lambda P)^{-1} X^{\top}y \\
        & = (X^{\top}X)^{+} \cdot V D_s (\Sigma+\lambda I)^{-1}(\Sigma+\lambda I)V^\top \cdot X^{\top}y \\
        &= (X^{\top}X)^{+} \tilde{V} \tilde{V}^\top X^{\top}y \\
        &= (X^{\top}X)^{+}X^{\top}y =X^{+}y,\\
    \end{align*}
where we use in the second line again a full SVD with $X=U\Sigma^\frac{1}{2} V^\top$ and the fact that diagonal matrices commute, in the third line a compact SVD with $X=\tilde{U}\tilde{\Sigma}^\frac{1}{2} \tilde{V}^\top$ and the fact that $V D_s V^\top = \tilde{V}\tilde{V}^\top$, and in the forth the fact that $\tilde{V}\tilde{V}^\top$, the projection onto the row space of $X$, can be disregarded as $X\top y$ is already in the row space. Finally, $X^{+}y$ denotes the well-known min-norm solution of the (underdetermined) least squares problem, i.e. when $n=r<p$.
\end{proof}

\subsection{Proof of \cref{thm:exp_reg_boost}}
\begin{proof}
From \eqref{eq:shrinkage_pspline} in \cref{prop_pspline_joint} we get an explicit expression for the parameters $\beta^{[k]}$ in each step $k$ of penalized $L_2$-Boosting (with quadratic penalty) and joint updates. With $\beta^{PLS} := (X^{\top}X + \lambda P)^{-1} X^{\top}y$ this is:
\begin{align*}
    \beta^{[k]} &= [\sum_{m=0}^{k-1} \nu (I - \nu (X^{\top}X + \lambda P)^{-1}X^{\top}X)^{m}] \ \beta^{PLS}\\
&= (X^{\top}X)^{-1}(X^{\top}X + \lambda P) \ [I - (I - \nu (X^{\top}X + \lambda P)^{-1}X^{\top}X)^{k}] \ \beta^{PLS}.
\end{align*}
In the second line, we used a telescope-sum argument.
In order to find the explicit minimization problem that is implicitly solved by boosting in each step, we can derive the explicit problem for which the solution of the problem corresponds to the above parameter formula in each step. For the derivation we build on ideas described in \cite{ali2019continuous}. Considering quadratic minimization problems, we aim to find the penalty matrix $\Gamma_k$ of the explicit minimization problem:
\begin{equation}{\label{q_problem}}
     \ell_{\Gamma_k}(\beta) = \frac{1}{2} \|y - X\beta\|^{2}+ \frac{1}{2} \beta^{\top}\Gamma_k\beta,
\end{equation}
for which the solution of the problem $\beta_{\Gamma_k}^{*}=(X^{\top}X + \Gamma_k)^{-1}X^{\top}y$ matches the above closed-form parameter expression of boosting (uniqueness of $\beta_{\Gamma_k}^{*}$ is guaranteed as $X$ has full column rank). To match them, we equate the two and get
\begin{align*}
    && (X^{\top}X + \Gamma_k)^{-1}X^{\top}y & \ = \ (X^{\top}X)^{-1}(X^{\top}X + \lambda P) \ [I - (I - \nu (X^{\top}X + \lambda P)^{-1}X^{\top}X)^{k}] \ (X^{\top}X + \lambda P)^{-1}X^{\top}y \\
    \Leftrightarrow && (X^{\top}X + \Gamma_k)^{-1} 
    &\ = \ (X^{\top}X)^{-1}(X^{\top}X + \lambda P) \ [I - (I - \nu (X^{\top}X + \lambda P)^{-1}X^{\top}X)^{k}] \ (X^{\top}X + \lambda P)^{-1}.
\end{align*}
By inverting both sides and subsequently rearranging terms, we obtain
$$\Gamma_k \ = \ - (X^{\top}X) + (X^{\top}X + \lambda P) \ [I - (I - \nu (X^{\top}X + \lambda P)^{-1}X^{\top}X)^{k}]^{-1} \ (X^{\top}X + \lambda P)^{-1} (X^{\top}X).$$
Again full column rank of $X$ ensures the invertibility of the terms. Using the Woodbury matrix identity, we can rewrite the inverse such that certain terms cancel out 
\begin{align*}
    \Gamma_k \ & = \ - (X^{\top}X) + (X^{\top}X + \lambda P) \ (I + [(I - \nu (X^{\top}X + \lambda P)^{-1}X^{\top}X)^{-k} - I]^{-1}) \ (X^{\top}X + \lambda P)^{-1} (X^{\top}X) \\
    & = \ (X^{\top}X + \lambda P) \ [ (I - \nu (X^{\top}X + \lambda P)^{-1}X^{\top}X)^{-k} - I]^{-1} \ (X^{\top}X + \lambda P)^{-1} (X^{\top}X)
\end{align*}
Finally, we use the notation $S_{\lambda} := (X^{\top}X + \lambda P)^{-1}X^{\top}X$ to obtain the desired result
$$\Gamma_k \ = \ (X^{\top}X) \ S_{\lambda}^{-1}  \ [ (I - \nu S_{\lambda})^{-k} - I]^{-1} \ S_{\lambda}.$$
\end{proof}

\subsection{Proof of \cref{matchridge}}{\label{app:matchridge}}
\begin{proof}
The connection of $L_2$-Boosting of linear models with joint updates to the solution paths of ridge regression under certain assumptions on the design matrix $X$ as well as the differences between the two in general, can be seen as a direct application of \cref{thm:exp_reg_boost}. The same holds true for boosting linear models with ridge penalty (ridge boosting). To see this, we can fist note that \cref{thm:exp_reg_boost} applies to linear model boosting and ridge boosting when choosing $\lambda=0$ and $P=I$, respectively. In order to link the solution paths of ridge regression to the paths of boosting, we need to link the explicit regularization term $\Gamma_k$ from \eqref{eq:exp_reg_boost} in \cref{thm:exp_reg_boost}, that is induced by the implicit minimization of boosting, to the penalization parameter $\tilde{\lambda}$ of ridge regression. We use different notations for the regularization terms in order to differentiate the explicit penalty in ridge regression from the penalty parameter $\lambda$ used in ridge boosting. We first derive the connection for linear model boosting and subsequently for ridge boosting. 

\paragraph{$L_2$-Boosting of Linear Models with Joint Updates} For $\lambda=0$ the quadratic penalty term $\Gamma_k$ in \eqref{eq:exp_reg_boost} simplifies:
\vspace*{-2mm}
\begin{equation*}
    \Gamma_k \ = \ (X^{\top}X) \ [(I - \nu I)^{-k} - I]^{-1}
    = \ (X^{\top}X) \ \frac{(1-\nu)^k}{1- (1-\nu)^k}.
\end{equation*}
Clearly, $\Gamma_k$ can now match the scalar penalty parameter of ridge regression $\tilde{\lambda}$, \textit{if and only if} $X^{\top}X=\sigma^2_X I$ for an arbitrary $\sigma^2>0$ ($\Gamma_k = \tilde{\lambda}(k) \cdot I \Leftrightarrow X^{\top}X=\sigma^2_X I$). With this condition on the design matrix, we get:
\begin{equation*}
    \Gamma_k = I \cdot \sigma_X^{2} \frac{(1-\nu)^k}{1- (1-\nu)^k} \quad \Leftrightarrow \quad \tilde{\lambda}(k) := \sigma_X^{2} \frac{(1-\nu)^k}{1- (1-\nu)^k}.
    \vspace*{-1mm}
\end{equation*}
Thus, with the condition on the design matrix, the parameters of $L_2$-Boosting of linear models with joint updates at iteration $k \in \mathbb{N}_0$ correspond to the solution of ridge regression with penalty parameter $\tilde{\lambda}(k)$.
\vspace*{-1mm}
\paragraph{Ridge Boosting with Joint Updates} For $P=I$ and the penalty parameter $\lambda>0$ of the ridge base learner, the quadratic penalty term $\Gamma_k$ in \eqref{eq:exp_reg_boost} simplifies to:
\begin{align*}
    \Gamma_k \ = \ (X^{\top}X + \lambda I) \ [(I - \nu (X^{\top}X + \lambda I)^{-1}X^{\top}X)^{-k} - I]^{-1} (X^{\top}X + \lambda I)^{-1}X^{\top}X.
\end{align*}
Again, $\Gamma_k$ can match the scalar penalty parameter of ridge regression $\tilde{\lambda}_{RB}$, \textit{if and only if} $X^{\top}X=\sigma^2_X I$ for an arbitrary $\sigma^2>0$ ($\Gamma_k = \tilde{\lambda}_{RB}(k) \cdot I \Leftrightarrow X^{\top}X=\sigma^2_X I$). With this condition on the design matrix, we get:
\begin{equation*}
    \Gamma_k = I \cdot \sigma_X^{2} \left[\left(1-\nu \frac{\sigma_X^{2}}{\sigma_X^{2} + \lambda}\right)^{-k}-1\right]^{-1} \quad \Leftrightarrow \quad \tilde{\lambda}_{RB}(k) := \sigma_X^{2} \left[\left(1-\nu \frac{\sigma_X^{2}}{\sigma_X^{2} + \lambda}\right)^{-k}-1\right]^{-1}.
\end{equation*}
Similar as before, the parameters of ridge boosting with joint updates at iteration $k \in \mathbb{N}_0$ correspond to the solution of ridge regression with penalty parameter $\tilde{\lambda}_{RB}(k)$ only under the condition of isotropic features ($X^{\top}X=\sigma^2_X I$). Otherwise, the implicit shrinkage of boosting for the two considered model classes cannot be matched to ridge regression in each step $k$ of the procedure. Lastly, we can notice that the two regularization parameters $\tilde{\lambda}(k)$ and $\tilde{\lambda}_{RB}(k)$ that characterize the implicit shrinkage of boosting explicitly, decrease in the boosting steps $k$ and converge to zero in the limit, in line with the results in \eqref{eq:linear_mod_boost} and \cref{prop_pspline_joint}. 
\end{proof}

\subsection{Derivation of \cref{prop:equi_gbcd}} \label{app:thm_equi_gbcd}

\subsubsection{Equivalence to GBCD with GSQ Update Scheme} 
\begin{derivation} The GSQ rule in block-wise $L_2$-Boosting for linear models can be recovered by examining the greedy selection of blocks at iteration $k$ (given the current residuals $u^{[k]}$) in the boosting method:
%
\begin{equation}{\label{selc_l2boost_gsq}}
\begin{aligned}
 \hat{b}_{k} & = \underset{b \in \mathcal{B}}{\arg \min } \left\|u^{[k]} - X_{b} \hat{\beta}_{b}\right\|^{2} \\
& = \underset{b \in \mathcal{B}}{\arg \min } \quad -  {u^{[k]}}^{\top}X_{b}\left(X_{b}^{\top}X_{b}\right)^{-1}X_{b}^{\top}u^{[k]}\\
& = \underset{b \in \mathcal{B}}{\arg \max } \quad \sqrt{(\nabla_{b}\ell(\beta^{[k]}))^\top \left(H_{b}\right)^{-1}\nabla_{b} \ell(\beta^{[k]})}\\
& = \underset{b \in \mathcal{B}}{\arg \max } \quad \| \nabla_{b}\ell(\beta^{[k]}) \|_{{H_{b}}^{-1}} \quad \text{(GSQ rule)}. 
\end{aligned}
\end{equation}
Here, $X_b$ corresponds to the $b$-th block of $X$, i.e., $b\in\mathcal{P}(\{1,\ldots,p\})$ with power set $\mathcal{P}$ with $\cup_{b\in\mathcal{B}} b = \{1,\ldots,p\}$ and $b_1\cap b_2 = \emptyset \,\forall b_1,b_2\in\mathcal{B}, b_1 \neq b_2$. After plugging in the OLS estimator for $\hat{\beta}_{b}$, we obtain the result by using the gradient and Hessian of the LS problem along the block $b$. Apart from recovering the GSQ rule \eqref{eq:gsq}, we further notice that the update step for this $L_2$-Boosting variant is 
\begin{equation} 
    \beta^{[k+1]} = \beta^{[k]} + \nu U_{\hat{b}_{k}} \hat{\beta}_{\hat{b}_{k}},
\end{equation}
with step size $\nu \in (0,1]$, $U_{\hat{b}_{k}}$ as defined in \cref{app:gbcd}, and $\hat{\beta}_{\hat{b}_{k}}$ as 
\begin{equation} 
    \hat{\beta}_{\hat{b}_{k}} = 
    \left(X_{\hat{b}_{k}}^{\top}X_
    {\hat{b}_{k}}\right)^{-1}
    X_{\hat{b}_{k}}^{\top}u^{[k]}
    = - (H_{\hat{b}_{k}})^{-1} \nabla_{\hat{b}_{k}} \ell\left(\beta^{[k]} \right).
\end{equation}

Thus, the update is identical to the GSQ update step as defined in \cref{app:updaterules} used for the GBCD routine as defined in \cref{app:gbcd}. Therefore, we have established the equivalence of block-wise $L_2$-Boosting and GBCD with GSQ-type selection and updates. 
\end{derivation}

\subsubsection{Equivalence to GSL Update Scheme} 
\begin{derivation}
Similarly, we can investigate block-wise $L_2$-Boosting for the case with only a single predictor per block. The greedy selection of components in $L_2$-Boosting at iteration $k$ (given the current residuals $u^{[k]}$) is
\begin{equation*}
    \begin{aligned}
 \hat{\jmath}_{k} & = \underset{1 \leq j \leq d}{\arg \min } \left\|u^{[k]} - X_{j} \hat{\beta}_{j}\right\|^{2} 
 = \underset{1 \leq j \leq d}{\arg \min } \, - \frac{\left\|X_{j}^{\top}u^{[k]}
 \right\|^{2}}{X_{j}^\top X_{j}}\\
& = \underset{1 \leq j \leq d}{\arg \max } \quad \frac{\left\|\nabla_{j} \ell(\beta^{[k]})\right\|^{2}}{L_{j}} \quad \text{(GSL rule).}
\end{aligned}
\end{equation*}
Here, $X_{j}$ corresponds to the column of $X$ belonging to the single predictor with index $j$. 
As for the block-wise selection, we plug in the OLS estimator for $\hat{\beta}_{b}$ that minimizes the OLS problem for each predictor. Using the coordinate-wise gradient and Hessian, we can recover the GSL selection rule \eqref{gsl-rule}. As before, we can examine the update step for this $L_2$-Boosting variant, which is 
\begin{equation} 
    \beta^{[k+1]} = \beta^{[k]} + \nu e_{\hat{\jmath}_{k}} \hat{\beta}_{\hat{\jmath}_{k}},
\end{equation}
with step size $\nu \in (0,1]$, $e_{\hat{\jmath}_{k}}$ as defined in \cref{app:gbcd}, and $\hat{\beta}_{\hat{\jmath}_{k}}$ as
\begin{equation} \label{comp_boost_update2}
    \hat{\beta}_{\hat{\jmath}_{k}} = \frac{ X_{\hat{\jmath}_{k}}^{\top}u^{[k]}
    }{X_{\hat{\jmath}_{k}}^{\top}X_{\hat{\jmath}_{k}}} = - \frac{\nabla_{\hat{\jmath}_{k}} \ell(\beta^{[k]})}{L_{\hat{\jmath}_{k}}}.
\end{equation}
This, again can be seen to be identical to the GSL update step as defined in \eqref{gsl-update}. Therefore, we have also established the equivalence of component-wise $L_2$-Boosting for linear models and GCD with GSL-type selection and updates.
\end{derivation}

\subsubsection{Derivation for \cref{rm:equi_gbcd}}\label{app:rm_equi_gbcd}

\begin{derivation} For $L_2$-Boosting with penalized linear models, we can also write the selection and update steps in terms of the gradient and Hessian. A key insight is that compared to G(B)CD applied to penalized problems, these BAM variants neglect the penalization accumulated in previous boosting iterations. We demonstrate this along the example of BAMs with block-wise regression spline fitting, which uses the penalized loss in \eqref{p_spline_problem}. In the derivation below each $X_b$ corresponds to the columns of a single regression spline, e.g., a P-spline. In this case, the greedy selection at step $k$ can be written as
\begin{equation}\label{selc-p-splines}
\begin{aligned}
 \hat{b}_{k} & = \underset{b \in \mathcal{B}}{\arg \min } \left\|u^{[k]} - X_{b} \hat{\beta}_{b}\right\|^{2}  + \lambda \hat{\beta}_{b}^{\top} P_{b} \hat{\beta}_{b}\\
& = \underset{b \in \mathcal{B}}{\arg \min } -  {u^{[k]}}^{\top}X_{b}\left(X_{b}^{\top}X_{b} + \lambda P_{b}\right)^{-1}{X}_{b}^{\top}u^{[k]}\\
& = \underset{b \in \mathcal{B}}{\arg \max } \| \nabla_{b}\ell_{LS}(\beta^{[k]}) \|_{(H^{PLS}_{b})^{-1}} \quad \text{(GSQ rule)}.
\end{aligned}
\end{equation}
In the second line in \eqref{selc-p-splines}, we plugged in the P-spline estimator for $\hat{\beta}_{b}$ and subsequently simplified terms. Importantly, in the fourth line, we recover the gradient of the unpenalized LS problem $\nabla_{b}\ell_{LS}$ as the algorithm does not account for penalization from previous steps. 
However, as the $H^{LS}_b \preceq H^{PLS}_b$, block-wise boosting with penalized $L_2$ loss can still be interpreted as GBCD with GSQ rule applied to the unpenalized $L_2$ loss.
The same can be observed for the update step which is
\begin{equation}{\label{p_splines_update1}}
    \beta^{[k+1]} = \beta^{[k]} + \nu 
    U_{\hat{b}_{k}} \hat{\beta}_{\hat{b}_{k}}
\end{equation}
with step size $\nu \in (0,1]$, $U_{\hat{b}_{k}}$ as defined in \cref{app:gbcd}, and $\hat{\beta}_{\hat{b}_{k}}$ as
\begin{equation}{\label{p_splines_update2}}
\begin{aligned}
    \hat{\beta}_{\hat{b}_{k}} 
    = & - \left(H^{PLS}_{\hat{b}_{k}}\right)^{-1} \ \nabla_{\hat{b}_{k}} \ell^{LS}(\beta^{k}).
\end{aligned}
\end{equation}
Note, that \eqref{selc-p-splines} and \eqref{p_splines_update2} scale the gradient with $H^{PLS}_b$, which is a block from the Hessian of the penalized problem. Hence, \eqref{selc-p-splines} is not equivalent to the update step that we have seen for $L_2$-Boosting \eqref{selc_l2boost_gsq}. Further, in case GBCD with GSQ rule is applied to the same penalized problem in \cref{p_spline_problem}, the selection and updates steps would be identical to \eqref{selc-p-splines} and \eqref{p_splines_update2}, with the important distinction that the gradient of the penalized problem is used. Hence, the two procedures would not be equivalent in this case.
\end{derivation}

\subsection{Proof of \cref{thm:block_conv}} \label{app:thm_block_conv}
\begin{proof}
In the following, we adopt some of the techniques used by \cite{nutini2022}, who showed convergence for GBCD with GSQ rule but without deriving an explicit convergence rate.
First, we use the fact that the function $\ell(\beta)$ is assumed to be $L$-smooth in the parameters $\beta \in \mathbb{R}^{p}$. This guarantees that there exists a matrix $H_b$ such that $\nabla_{bb}^{2} \ell(\beta) \preceq H_b$ $\forall \beta \in \mathbb{R}^p$, $\forall b \in \mathcal{B}$. For the considered problems, this also guarantees that there exists $\nu \in (0,1]$ s.t.\ $\nabla_{bb}^{2} \ell(\beta) \preceq \frac{1}{\nu}X_{b}^{\top}X_{b}$ $\forall \beta \in \mathbb{R}^p$, $\forall b \in \mathcal{B}$. In the following, the matrix $H_b$ may therefore denote $\frac{1}{\nu}X_{b}^{\top}X_{b}$ or $\frac{1}{\nu}(X_{b}^{\top}X_{b} + \lambda P_b)$ for unpenalized and penalized base learners, respectively. The latter also provides a valid upper bound of the Hessian in case the former does, given that $0\preceq\lambda P_b$. Using the matrix formulation of the $L$-smoothness condition \eqref{block-lip}, we can derive the following upper bound: 
\begin{equation}{\label{eq:progbound}}
\begin{aligned}
    \ell(\beta^{[k+1]}) & \leq \ell(\beta^{[k]}) + \langle \nabla_{b_{k}} \ell(\beta^{[k]}), \beta^{[k+1]} - \beta^{[k]} \rangle + \frac{1}{2} \|\beta^{[k+1]} - \beta^{[k]}\|_{H_{b_{k}}}^{2} \\
    & = \ell(\beta^{[k]}) - \frac{\nu}{2} \|\nabla_{b_{k}} \ell(\beta^{[k]})\|_{{H_{b_{k}}}^{-1}}^{2}\, ,
\end{aligned}
\end{equation}
where the first inequality follows from the ${L}$-smoothness in \eqref{block-lip} and for the second equality we used the GSQ related update \eqref{block_update_2} for $\beta^{[k+1]}$. 
%
We can rewrite the upper bound in \eqref{eq:progbound} by using the norm
\begin{equation}{\label{b_norm}}
    \|\vartheta\|_{\mathcal{B}} = \max_{b\in\mathcal{B}} \|\vartheta_{b}\|_{{H_{b}}^{-1}}
\end{equation}
for some $\vartheta \in \mathbb{R}^{p}, \vartheta_b\in\mathbb{R}^{|b|}$, and some p.d.\ $H_{b} \in \mathbb{R}^{|b|\times|b|}$. 
Using this norm, we have that \mbox{$\|\nabla \ell(\beta^{[k]})\|_{\mathcal{B}} = \|\nabla_{b_{k}} \ell(\beta^{[k]})\|_{{H_{b_{k}}}^{-1}}$}, so that we can write \eqref{eq:progbound} as
\begin{equation}
\label{eq:prog_bound_b_norm}
    \ell(\beta^{[k+1]}) \leq \ell(\beta^{[k]}) - \frac{\nu}{2} \|\nabla \ell(\beta^{[k]})\|_{\mathcal{B}}^2.
\end{equation}
Next we use the fact that the function $\ell(\beta)$ is assumed to be $\mu$-PL in the parameters $\beta \in \mathbb{R}^{p}$. Instead of using \cref{def:PL} in terms of the $L_2$-norm, we use the previously introduced norm in \eqref{b_norm}, for which it holds $\|\vartheta\|_{2}^{2}\leq L_{\mathcal{B}} |\mathcal{B}| \|\vartheta\|_{\mathcal{B}}^{2}$ with $L_{\mathcal{B}} = \max_{b\in \mathcal{B}} \lambda_{max}(H_{b})$. This inequality can be verified by
\begin{equation}
    \|\vartheta\|^{2}_{2} = \sum_{b}^{|\mathcal{B}|} \|\vartheta_{b}\|^{2}_{2} \leq \sum_{b}^{|\mathcal{B}|} \lambda_{max}(H_{b}) \ \vartheta_{b}^{\top}{H_{b}}^{-1}\vartheta_{b} \leq L_{\mathcal{B}} \sum_{b}^{|\mathcal{B}|} \vartheta_{b}^{\top}{H_{b}}^{-1}\vartheta_{b} \leq L_{\mathcal{B}} \ |\mathcal{B}| \|\vartheta\|^{2}_{\mathcal{B}}.
\end{equation} 
Thus, we obtain the PL-inequality
\begin{equation}{\label{pl_ineq_bnorm}}
    \frac{1}{2} \| \nabla \ell(\beta) \|_{\mathcal{B}}^{2} \: \geq \: \frac{\mu}{L_{\mathcal{B}} |\mathcal{B}|} \: (\ell(\beta) - \ell^{*}),
\end{equation}
where $\mu$ corresponds to the PL parameter of \cref{def:PL} with $L_{2}$-norm.
Lastly, by connecting the inequalities \eqref{eq:prog_bound_b_norm} and \eqref{pl_ineq_bnorm}, iterating over $k$ iterations, and subtracting $\ell^*$ from both sides, we get our final convergence result
\begin{equation} \label{eq:conv_bgcd}
    \ell(\beta^{[k]})-\ell^{*} \leq \left(1- \nu \frac{\mu}{L_{\mathcal{B}} |\mathcal{B}|} \right)^{k}\left(\ell(\beta^{[0]})-\ell^{*}\right).
\end{equation}
\end{proof}

\subsection{Proof of \cref{corr:quad_prob}} \label{app:corr_block_conv}
\begin{proof}
In case of quadratic problems, the Hessian upper bound condition ($\nabla_{bb}^{2} \ell(\beta) \preceq H_b$ $\forall \beta \in \mathbb{R}^p$, $\forall b \in \mathcal{B}$) from \cref{thm:block_conv} is fulfilled for any $\nu \in (0,1]$, thus, in particular for $\nu = 1$. In this case, it suffices to consider the bound in \eqref{eq:progbound} with $H_b$ denoting $X_{b}^{\top}X_{b}$ or $X_{b}^{\top}X_{b} + \lambda P_b$ for unpenalized and penalized base learners, respectively. Therefore, we get the slightly refined bound
\begin{equation}{\label{eq:progbound_quad}}
\begin{aligned}
    \ell(\beta^{[k+1]}) & \leq \ell(\beta^{[k]}) + \langle \nabla_{b_{k}} \ell(\beta^{[k]}), \beta^{[k+1]} - \beta^{[k]} \rangle + \frac{1}{2} \|\beta^{[k+1]} - \beta^{[k]}\|_{H_{b_{k}}}^{2} \\
    & = \ell(\beta^{[k]}) - \nu(1 - \frac{\nu}{2}) \|\nabla_{b_{k}} \ell(\beta^{[k]})\|_{{H_{b_{k}}}^{-1}}^{2}\, ,
\end{aligned}
\end{equation}
where again the first inequality follows from the ${L}$-smoothness in \eqref{block-lip} and for the second equality we used the GSQ related update \eqref{block_update_2} for $\beta^{[k+1]}$. 
Similar to \eqref{eq:prog_bound_b_norm}, by using again the norm in \eqref{b_norm}, we can write \eqref{eq:progbound_quad} as
\begin{equation}
\label{eq:prog_bound_b_norm_quad}
    \ell(\beta^{[k+1]}) \leq \ell(\beta^{[k]}) - \nu(1 - \frac{\nu}{2}) \|\nabla \ell(\beta^{[k]})\|_{\mathcal{B}}^2.
\end{equation}
Analogously to \eqref{eq:conv_bgcd}, by connecting the inequalities \eqref{pl_ineq_bnorm} and \eqref{eq:prog_bound_b_norm_quad}, iterating over $k$ iterations, and subtracting $\ell^*$ from both sides, we get the following convergence result (with slightly faster convergence rate)
\begin{equation} 
    \ell(\beta^{[k]})-\ell^{*} \leq \left(1- \nu (2 - \nu) \frac{\mu}{L_{\mathcal{B}} |\mathcal{B}|} \right)^{k}\left(\ell(\beta^{[0]})-\ell^{*}\right).
\end{equation}
We can make the convergence rate more explicit, noting that for quadratic functions (with Hessian $Q$) in particular, it holds that $\mu=\lambda_{pmin}(Q)$. Further we can use that $L_{\mathcal{B}}\leq\lambda_{max}(Q)$ to get
\begin{equation} 
    \ell(\beta^{[k]})-\ell^{*} \leq {\left(1- \nu (2 - \nu) \frac{1}{|\mathcal{B}|} \frac{\lambda_{pmin}(Q)}{\lambda_{max}(Q)} \right)}^{k}\left(\ell(\beta^{[0]})-\ell^{*}\right).
\end{equation}
\end{proof}

\subsection{Proof of \cref{corr:conv_regspline}} \label{app:conv_regspline}
\begin{proof}

\cref{corr:conv_regspline} essentially follows from \cref{thm:block_conv}. This, is because the proof of \cref{thm:block_conv} in \cref{app:thm_block_conv} can be conducted analogously for GBCD with GSQ. The only difference compared to block-wise boosting is that GBCD with GSQ is using gradients of the penalized problem in the selection and update steps (see \cref{app:rm_equi_gbcd}). Therefore, under the assumptions of \cref{corr:conv_regspline}, i.e., that the regression spline is $\mu$-PL and $L$-smooth in its parameters,  \cref{thm:block_conv} implies convergence to a solution of the regression spline with minimal loss. The two conditions of $\mu$-PL and $L$-smoothness follow from the fact that regression spline problems can be written in quadratic form (cf.~\cref{app:funprop}). Note, in case the regression spline problem is not only $\mu$-PL but strongly convex, GBCD with GSQ converges to the unique optimal solution.
\end{proof}

\subsection{Derivation of \cref{prop:css}} \label{app:multiple_csp}

\begin{derivation}
Assume $d:=|\mathcal{B}|$ cubic smoothing splines, each with operator $S_{m}:\mathbb{R}^{n} \to \mathbb{R}^{n}, (m\in [d])$, that maps the current residuals to fitted values. Each $S_{m}$ has $n$ eigenvalues $(\lambda_m)_{i}, i\in [n]$, with $0<(\lambda_m)_{i} \leq 1 \,\forall i\in [n]$. Define the boosting operator in step $k$ as $T^{[k]}:=\left(I-S_{m_{k}}\right)$, where $m_{k}$ denotes the index of the selected cubic smoothing spline learner at step $k$. The boosting operator $T^{[k]}:\mathbb{R}^{n} \to \mathbb{R}^{n}$ with $k \in \mathbb{N}_0$ has $n$ eigenvalues $(\tilde{\lambda}_m)_{i}$, $i\in [n]$, with $0 \leqslant (\tilde{\lambda}_m)_{i}<1$ \citep{BuehlmannYu2003}. Define $\tilde{\lambda}_{max} := \max_{m \in [d]} ( \max_{i \in [n]} ((\tilde{\lambda}_m)_{i}))$, to be the largest eigenvalue of all $d$ boosting operators. Further, the fitted values $f^{\ast}$ of the saturated model match the observed values $y$ exactly, i.e., $f^{\ast}=y$. Then it holds
$$
\begin{aligned}
\left\|y-f^{[k]}\right\|= \left\|f^{\ast}-f^{[k]}\right\|&=\| T^{[k]} \cdot \ldots \cdot T^{[1]} y \|\\
& \leq \|T^{[k]} \cdot \ldots \cdot T^{[1]}\left\|^{*}\right\| y \| &\text{by Property \ref{proper:op}, \ref{pf:css_pre2}.} \\
& \leq\left\|T^{[k]}\right\|^{*} \cdot \ldots \cdot\left\|T^{[1]}\right\|^{*}\|y\| &\text{by Property \ref{proper:op}, \ref{pf:css_pre3}.}\\
& \leq \|y\| \underbrace{( \underbrace{\tilde{\lambda}_{max}}_{\in [0,1)})^{k}}_{\stackrel{k \rightarrow \infty}{\rightarrow} 0}
\end{aligned}
$$
Thereby, it follows that $f^{[k]} \rightarrow f^\ast$ for $k \rightarrow \infty$.   
\end{derivation}

\subsection{Derivation of \cref{prop:expo_conv} and Further Remarks}\label{app:expfam_gbcd}
\begin{derivation}
As described in Algorithm \ref{algo:GB}, in each iteration BAMs separately fit each base learner against the negative functional derivative of the loss function $\ell(\cdot)$ at the current function estimate $f^{[k]}$. 
In order to relate boosting for exponential families to GBCD, we establish the link between the negative functional derivative $\{\widetilde{y_{i}}\}^{n}_{i=1}$ and the gradient of the loss function with respect to the parameter $\beta$. Considering the negative log-likelihood instead of the log-likelihood in \eqref{eq:log_lik_expfam} and using the linearity of the base learners in their parameters ($f = X\beta$), one can do so by
\begin{equation}
      -\frac{\partial}{\partial \beta} \ell(\beta) =  \frac{\partial}{\partial \beta}  f(\beta) \: \: \widetilde{y} = X^{\top}\widetilde{y}.
\end{equation}
The block-wise LS base procedure of BAMs for exponential family loss can now be written as GBCD with GSQ. We derive this for the block-wise procedure. The component-wise procedure then follows as a special case. First, the block selection is done via
\begin{equation}{\label{selc_boost_expfam}}
\begin{aligned}
 \hat{b}_{k} & = \underset{b \in \mathcal{B}}{\arg \min } \left\|\widetilde{y}^{[k]} - X_{b} \hat{\beta}_{b}\right\|^{2} \\
& = \underset{b \in \mathcal{B}}{\arg \min } \quad -  \widetilde{y}^{[k]^{\top}} X_{b}\left(X_{b}^{\top}X_{b}\right)^{-1}{X}_{b}^{\top}{\widetilde{y}}^{[k]}\\
& = \underset{b \in \mathcal{B}}{\arg \max } \quad \sqrt{(\nabla_{b}\ell(\beta^{[k]}))^\top \left(X_{b}^{\top}X_{b}\right)^{-1}\nabla_{b} \ell(\beta^{[k]})}\\
& = \underset{b \in \mathcal{B}}{\arg \max } \quad \| \nabla_{b}\ell(\beta^{[k]}) \|_{(X_{b}^{\top}X_{b})^{-1}} \quad \text{(GSQ rule)},
\end{aligned}
\end{equation}
which is analogous to the derivation in \eqref{selc_l2boost_gsq} and corresponds to the GSQ rule. Similarly, the update is
\begin{equation}
    \hat{\beta}^{[k+1]} = \hat{\beta}^{[k]} + \nu 
    U_{\hat{b}_{k}} \hat{\beta}_{\hat{b}_{k}}
\end{equation}
with
\begin{equation}
    \hat{\beta}_{\hat{b}_{k}} = \left(X_{\hat{b}_{k}}^{\top}X_{\hat{b}_{k}}\right)^{-1} X^{\top}_{\hat{b}_{k}}\widetilde{y}^{[k]}
    = - \left(X_{\hat{b}_{k}}^{\top}X_{\hat{b}_{k}}\right)^{-1} \ \nabla_{\hat{b}_{k}} \ell(\beta^{[k]}).
\end{equation}
Therefore, as long as $\frac{1}{\nu} X_{b}^{\top}X_{b}$ provides an upper bound to the respective blocks of the Hessian in \eqref{eq:hessian_expfam} for all $b \in \mathcal{B}$, the selection and updates correspond to the GSQ rule and we get linear convergence of BAMs with block-wise updates for exponential family loss due to \cref{thm:block_conv}. Note that the derivation essentially builds on the linearity of the base learners in their parameters. Thus, the same could be done for any other model class that fulfills this condition.
\end{derivation}

\subsection{Remark on other Loss Functions} \label{sec:remarkloss}

Robust loss functions such as the $L_{1}$ and Huber loss are frequently used alternatives to the $L_{2}$ loss for gradient boosting methods \citep{BuehlmannHothorn2007}.
While the idea of greedy selection and updates would still be applicable in this case, we cannot obtain convergence results as stated in \cref{thm:block_conv}. The reason for this is that the gradient of the $L_{1}$-loss does not decrease in size as we reach the minimizer of the problem, which is why it cannot be $\mu$-PL. The Huber loss is $\mu$-PL only in a $\delta$-neighborhood of the minimum. Thus, the convergence rate for the Huber loss cannot be globally linear.

Exponential family loss functions with canonical link such as the Binomial and Poisson loss were shown to be strictly convex in \cref{app:expfam} and thus $\mu$-PL over any compact set. While the Binomial loss was also shown to be $L$-smooth ($L=1/4$), the Poisson loss is non-globally Lipschitz continuous.
However, the Poisson loss is still L-smooth over any compact set (but potentially with a large Lipschitz constant $L$>0).

Lastly, we consider the loss of the Cox Proportional Hazards (Cox PH) model that is frequently encountered in survival analysis. For the Cox PH model, the log partial likelihood is given by
$$\ell(\beta) = \sum_{i : C_i = 1} \left( x_i^\top \beta - \log \sum_{j \in \tilde{R} (t_i)} \exp(x_j^\top \beta)\right)
$$
and the gradient is given by
$$\nabla \ell(\beta) = \sum_{i : C_i = 1} \left( x_i - \frac{\sum_{j \in \tilde{R}(t_i)} x_j \exp(x_j^\top \beta)}{\sum_{j \in \tilde{R}(t_i)} \exp(x_j^\top \beta)} \right) := \sum_{i : C_i = 1} \left( x_i - \sum_{j \in \tilde{R}(t_i)} x_j \omega_j(\beta)\right),
$$
where $t_i$ the observed survival time, $C_i$ is a event indicator ($C_i=1$ if the event occurred and $C_i=0$ if the data is censored), and $\tilde{R}(t_i)$ the risk set at time $t_i$ (the set of individuals still at risk of the event at time $t_i$). First, we consider the $\mu$-PL condition. Importantly, while the first term in the log partial likelihood is simply linear in $\beta$ the second is a log-sum-exp function of the parameters $\beta$. Apart from pathological cases that can be ruled out in almost all modeling setups, such functions are known to be strictly convex. Thus, similar to the case of exponential family losses, the $\mu$-PL condition is fulfilled for the Cox PH loss over any compact set. To show $L$-smoothness of the Cox PH loss, we consider the gradient of the loss, which only depends on $\beta$ via the term $\omega_j(\beta)$. As the latter can be identified as a softmax function, which is known to be Lipschitz continuous, the $L$-smoothness of the Cox PH loss follows. Given the two fulfilled conditions, convergence guarantees from \cref{thm:block_conv} apply. We demonstrate this using numerical experiments in \cref{app:surv_boost_exp}.

\subsection{Proof of \cref{prop:biconvex}}\label{app:biconvex}
\begin{proof}
    We now prove that the distributional Gaussian regression is biconvex in $(\beta,\xi)$. Given the negative log-likelihood \(-\mathcal{L}\) of a Gaussian distribution with the mean \(\psi_i = x_i^\top \beta\) and standard deviation \(\sigma_i = \exp(z_i^\top \xi)\), and observed values \(y_i\) for \(i=1,\ldots,n\), we first derive the gradient and Hessian with respect to \(\beta\) and \(\xi\). Note that $x_i$ and $z_i$ could contain the same or different features, and are thus not necessarily the same. The probability density function (PDF) of a Gaussian distribution is given by:
\[ p_{\mathcal{N}(\psi_i, \sigma_i)}(y_i) = \frac{1}{\sqrt{2\pi\sigma_i^2}} \exp\left( -\frac{(y_i - \psi_i)^2}{2\sigma_i^2} \right). \]
The log-likelihood for \(n\) observations is:
\[ \mathcal{L} = \sum_{i=1}^n \log p_{\mathcal{N}(\psi_i, \sigma_i)}(y_i). \]
Substituting the Gaussian PDF, we get:
\[ \mathcal{L} = \sum_{i=1}^n \left[ -\frac{1}{2} \log(2\pi) - \frac{1}{2} \log(\sigma_i^2) - \frac{(y_i - \psi_i)^2}{2\sigma_i^2} \right]. \]
Simplifying, and noting that \(\sigma_i = \exp(z_i^\top \xi)\) implies \(\log(\sigma_i^2) = 2z_i^\top \xi\), we get:
\[ -\mathcal{L} = \frac{n}{2} \log(2\pi) + \sum_{i=1}^n \left( z_i^\top \xi + \frac{(y_i - x_i^\top \beta)^2}{2 \exp(2z_i^\top \xi)} \right). \]
Let \(r_i = y_i - x_i^\top \beta\) and \(\sigma_i = \exp(z_i^\top \xi)\). The derivative with respect to \(\beta\) is
\[ \frac{\partial (-\mathcal{L})}{\partial \beta} = \sum_{i=1}^n \frac{\partial}{\partial \beta} \left( \frac{r_i^2}{2\sigma_i^2} \right)  = \sum_{i=1}^n \left( \frac{1}{2\sigma_i^2} \right) \left( -2 x_i r_i \right) = - \sum_{i=1}^n \frac{x_i r_i}{\sigma_i^2}. \]
The derivative with respect to \(\xi\) is
\[ \frac{\partial (-\mathcal{L})}{\partial \xi} = \sum_{i=1}^n \left( \frac{\partial z_i^\top \xi}{\partial \xi} + \frac{\partial}{\partial \xi} \left( \frac{r_i^2}{2\sigma_i^2} \right) \right). \]
The first term is \(\frac{\partial z_i^\top \xi}{\partial \xi} = z_i\), whereas the second term evaluates to
\[ \frac{\partial}{\partial \xi} \left( \frac{r_i^2}{2\sigma_i^2} \right) = \frac{\partial}{\partial \xi} \left( \frac{r_i^2}{2 \exp(2z_i^\top \xi)} \right) = -\frac{r_i^2}{2\exp(2z_i^\top \xi)} \cdot \frac{\partial \exp(2z_i^\top \xi)}{\partial \xi}. \]
As 
\[ \frac{\partial \exp(2z_i^\top \xi)}{\partial \xi} = 2 z_i \exp(2z_i^\top \xi). \]
we have
\[ \frac{\partial}{\partial \xi} \left( \frac{r_i^2}{2\sigma_i^2} \right) = - 2 z_i \cdot \frac{r_i^2}{2\sigma_i^2} = - z_i \frac{r_i^2}{\sigma_i^2}. \]
So, combining both terms:
\[ \frac{\partial (-\mathcal{L})}{\partial \xi} = \sum_{i=1}^n \left( z_i - z_i \frac{r_i^2}{\sigma_i^2} \right) = \sum_{i=1}^n z_i \left( 1 - \frac{r_i^2}{\sigma_i^2} \right). \]
For the Hessian we need to compute the second derivatives with respect to \(\beta\) and \(\xi\). We start with $\beta$:
\[ \frac{\partial^2 (-\mathcal{L})}{\partial \beta \partial \beta^\top} = \sum_{i=1}^n \frac{\partial}{\partial \beta^\top} \left( - \frac{x_i r_i}{\sigma_i^2} \right) = \sum_{i=1}^n \frac{x_i x_i^\top}{\sigma_i^2}. \]
For $\xi$ we have:
\[ \frac{\partial^2 (-\mathcal{L})}{\partial \xi \partial \xi^\top} = \sum_{i=1}^n \frac{\partial}{\partial \xi^\top} \left( z_i - z_i \frac{r_i^2}{\sigma_i^2} \right). \]
\[ \frac{\partial}{\partial \xi^\top} \left( z_i - z_i \frac{r_i^2}{\sigma_i^2} \right) = \frac{\partial z_i}{\partial \xi^\top} - \frac{\partial z_i}{\partial \xi^\top} \frac{r_i^2}{\sigma_i^2} - z_i \frac{\partial}{\partial \xi^\top} \left( \frac{r_i^2}{\sigma_i^2} \right). \]
Since \(\frac{\partial z_i}{\partial \xi^\top} = 0\) and noting that:
\[ \frac{\partial}{\partial \xi^\top} \left( \frac{r_i^2}{\sigma_i^2} \right) = -2 \frac{r_i^2}{\sigma_i^2} z_i^\top, \]
we get
\[ \frac{\partial^2 (-\mathcal{L})}{\partial \xi \partial \xi^\top} = \sum_{i=1}^n \left( 2 \frac{r_i^2}{\sigma_i^2} z_i z_i^\top \right). \]
For the mixed second derivatives with respect to \(\beta\) and \(\xi\), we have
\[ \frac{\partial^2 (-\mathcal{L})}{\partial \beta \partial \xi^\top} = \sum_{i=1}^n \frac{\partial}{\partial \xi^\top} \left( - \frac{x_i r_i}{\sigma_i^2} \right), \]
where
\[ \frac{\partial}{\partial \xi^\top} \left( - \frac{x_i r_i}{\sigma_i^2} \right) = - x_i r_i \frac{\partial}{\partial \xi^\top} \left( \frac{1}{\sigma_i^2} \right) = 2 \frac{r_i}{\sigma_i^2} \cdot x_i z_i^\top. \]
So the mixed Hessian component is:
\[ \frac{\partial^2 (-\mathcal{L})}{\partial \beta \partial \xi^\top} = 2 \sum_{i=1}^n \frac{r_i}{\sigma_i^2} \cdot x_i z_i^\top. \]
Combining everything, the Hessian matrix thus is
\[
H = \begin{bmatrix}
\frac{\partial^2 (-\mathcal{L})}{\partial \beta \partial \beta^\top} & \frac{\partial^2 (-\mathcal{L})}{\partial \beta \partial \xi^\top} \\
\frac{\partial^2 (-\mathcal{L})}{\partial \xi \partial \beta^\top} & \frac{\partial^2 (-\mathcal{L})}{\partial \xi \partial \xi^\top}
\end{bmatrix} = \begin{bmatrix}
\sum_{i=1}^n \frac{x_i x_i^\top}{\exp(2z_i^\top \xi)} & 2 \sum_{i=1}^n \frac{(y_i - x_i^\top \beta) x_i z_i^\top}{\exp(2z_i^\top \xi)} \\
2 \sum_{i=1}^n \frac{(y_i - x_i^\top \beta) z_i x_i^\top}{\exp(2z_i^\top \xi)} & 2 \sum_{i=1}^n \frac{(y_i - x_i^\top \beta)^2 z_i z_i^\top}{\exp(2z_i^\top \xi)}
\end{bmatrix}.
\]
We can directly observe that for given $\hat{\beta}$, the lower diagonal block 
can be written as ${Z}^\top {\Psi} {Z}$ with ${\Psi} = \text{diag}(2r_i^2 /\sigma_i^2)$ and ${Z} = (z_1, \ldots, z_n)^\top$, i.e., a positive semi-definite matrix. Clearly, due to the term $\exp(2z_i^\top \xi)$ on the diagonal, ${Z}^\top {\Psi} {Z}$ cannot be bounded for all $\xi$. Accordingly, \eqref{eq:loglik_gaussdist} cannot be $L$-smooth in the parameters $\xi$. The upper diagonal block of $H$ is a positive definite matrix since it can similarly be written as ${X}^\top {\Sigma}^{-1} {X}$ with ${\Sigma} = \text{diag}(\sigma_i^2)$. In general, however, the Hessian cannot be written as 
\[
H = \Lambda^\top {\Sigma}^{-1} \Lambda
\]
since with $\Lambda = (X \,\, \sqrt{2}\tilde{Z})$ and $\tilde{Z} = Z \text{diag}(r_i)$ the factors $2$ in front of the mixed terms cannot be matched. Hence, even if the diagonal blocks are positive definite, we can construct a counterexample, e.g., $n=1$, $x_i = z_i = \beta = \xi = 1$ and $y_i=2$, for which the matrix is indefinite.
\end{proof}

\newpage
\section{CUBIC SMOOTHING SPLINES AND P-SPLINES}\label{app:css_psplines}
\textbf{Cubic Smoothing Splines.} A \textit{cubic smoothing spline} $f$ as considered in Section~\ref{sec:css_regspline} (for a twice differentiable function $f$), minimizes
\begin{equation}{\label{cubic_smoothing_spline}}
    \sum^{n}_{i=1} (y_{i} - f(x_{i}))^{2} + \lambda \int \left(\frac{\partial^{2} f}{\partial x^{2}}\right)^{2} dx,
\end{equation}
for a given $\lambda \geq 0$ and where $x_{i}$ denotes the $i$-th row of the $n \times p$ feature matrix $X$. Cubic smoothing splines aim to find an optimal function that best adjusts to the least squares problem while at the same time penalizing the squared second derivative of the function to induce smoothness on the solution. The trade-off between the two is controlled by the smoothness parameter $\lambda$. Cubic smoothing splines are considered by \cite{BuehlmannYu2003} for component-wise $L_{2}$-Boosting. 

\textbf{P-splines} Although cubic smoothing splines are particularly interesting from a theoretical point of view, the integral for penalization imposes major computational disadvantages compared to other smoothing base learners. A natural alternative is to consider a discretized version of it. As such, \cite{eilers1996} proposed a specific type of penalized regression splines, which uses B-splines \citep{deBoor1978} as basis expansions of the predictor variables and penalizes the higher-order differences between adjacent regression parameters of the B-spline. 
B-splines, also called basis splines, are a collection of piece-wise polynomials that are connected at specific points, called knots. For notational convenience, we elaborate on the basis expansion of a single variable via B-splines. The extension to multiple variables is straightforward. Consider $\{B^{l}_{i}(\cdot)\}^{\kappa+l-1}_{i=1}$ B-spline basis functions of order $l$ defined at $\kappa$ equidistant knot positions. When considering a P-spline for a certain feature $Z_{j}$ $(j\in [p])$, this will induce an expanded feature matrix ${X}$ \citep{fahrmeir2013}, which is given by
$$ {X} = 
  \begin{pmatrix}
    B^{l}_{1}(Z_{1,j}) & \dots & B^{l}_{\kappa+l-1}(Z_{1,j}) \\
    \vdots &  & \vdots \\
    B^{l}_{1}(Z_{n,j}) & \dots & B^{l}_{\kappa+l-1}(Z_{n,j})
  \end{pmatrix}
.$$
With this expanded feature matrix, the approximation of the cubic smoothing spline by a P-spline is given by
\begin{equation}{\label{p_spline}}
    \| y - {X}\beta\|^{2} + \lambda \beta^{\top}\underbrace{ {D}_{2}^{\top} {D}_{2}}_{:={P}}\beta,
\end{equation}
where $\beta$ now denotes an extended parameter vector corresponding to each piecewise polynomial in the B-spline, respectively. The matrix ${P}$ corresponds to the second-order difference penalty matrix. The $p-2 \times p$ matrix ${D}_{2}$ and $p \times p$ matrix ${P}$ are defined as
\[
 {D}_{2} = 
\begin{pmatrix}
  1 & -2 & 1      &        &        & \\
    &  1 & -2     & 1      &        &\\
    &    & \ddots & \ddots & \ddots &\\
    &    &        & 1      & -2     & 1
\end{pmatrix}, \quad \quad \quad
{P} = 
\begin{pmatrix}
  1 & -2    & 1      &        &        &        & \\
 -2 &     5 & -4     & 1      &        &        &\\
  1 &    -4 & 6      & -4     &      1 &        &\\
    & \ddots& \ddots & \ddots & \ddots & \ddots &\\
    &       &      1 & -4     & 6      & -4     &1\\
    &       &        & 1      & -4     & 5      &-2\\
    &       &        &        & 1      & -2     &1
\end{pmatrix}.
\]
Note that the penalty matrix ${P}$ is rank deficient. 
A key property of the P-spline modeling approach in \eqref{p_spline} is that the dimensionality of the penalization term is greatly reduced compared to the penalty in \eqref{cubic_smoothing_spline}, as we consider a discretized version of it \citep{schmid2008}. This is what gives P-splines a great computational advantage over cubic smoothing splines and thus makes it particularly interesting for applications such as $L_{2}$-Boosting, where \eqref{p_spline} needs to be solved repeatedly.

\section{COMPARISON OF BAM WITH RELATED METHODS} \label{app:comparison_bams}

To emphasis the relevance of BAMs and our derived theoretical results,
we compare BAMs to closely related methods and briefly remark on related theoretical convergence guarantees for such model classes. In particular, we compare BAMs with Generalized Additive Models (GAMs), Neural Additive Models (NAMs), and Gradient Boosting Machines (GBMs).

\textbf{GAMs.}\quad Generalized Additive Models (GAMs) \citep{hastie1986generalized, wood2017} are a well-studied model class for which convergence guarantees, the limiting solution, and statistical inference results are well established. However, this model class usually requires setting several modeling aspects up front, e.g. additive terms to be included in the final model and the smoothness of each additive component (e.g. degree of a spline). BAMs can use the same additive components and incorporate the same link functions as GAMs (see \cref{sec:glmboosting}). However, BAMs are more flexible than GAMs by allowing for variable or base learner selection as well as an adaptive smoothness.

\textbf{NAMs.}\quad Neural Additive Models (NAMs) \citep{agarwal2021neural, radenovic2022neural} have recently sparked a lot of interest in the ML community as NAMs offer a very flexible modeling class by learning adaptive basis functions and smoothness through neural networks while still remaining interpretable due to their additive structure. While they share the interpretability property with BAMs, no explicit theoretical results on convergence or solution paths exist for NAMs, making uncertainty quantification and other inferential tasks challenging. Apart from difficulties in the optimization of NAMs, overparametrization, non-identifiability issues, and complex loss landscapes further complicate the derivation of any theoretical guarantees.

\textbf{GBMs.}\quad Gradient Boosting Machines (GBMs) \citep{friedman2001} also build on the idea of functional gradient descent, and when considering the class of linear or spline base learners, GBMs can be considered to be equivalent to BAMs. While GBMs are often used in combination with trees, very few theoretical results exist about convergence guarantees and the solution paths of such variants, given that conditions such as $\mu$-PL do not necessarily need to hold, when considering trees as base learners.

\section{ADDITIONAL EXPERIMENTS, EXPERIMENTAL DETAILS AND RESULTS} \label{app:further}

\begin{figure}[ht!]
\begin{center}
    \includegraphics[width=0.7\columnwidth]{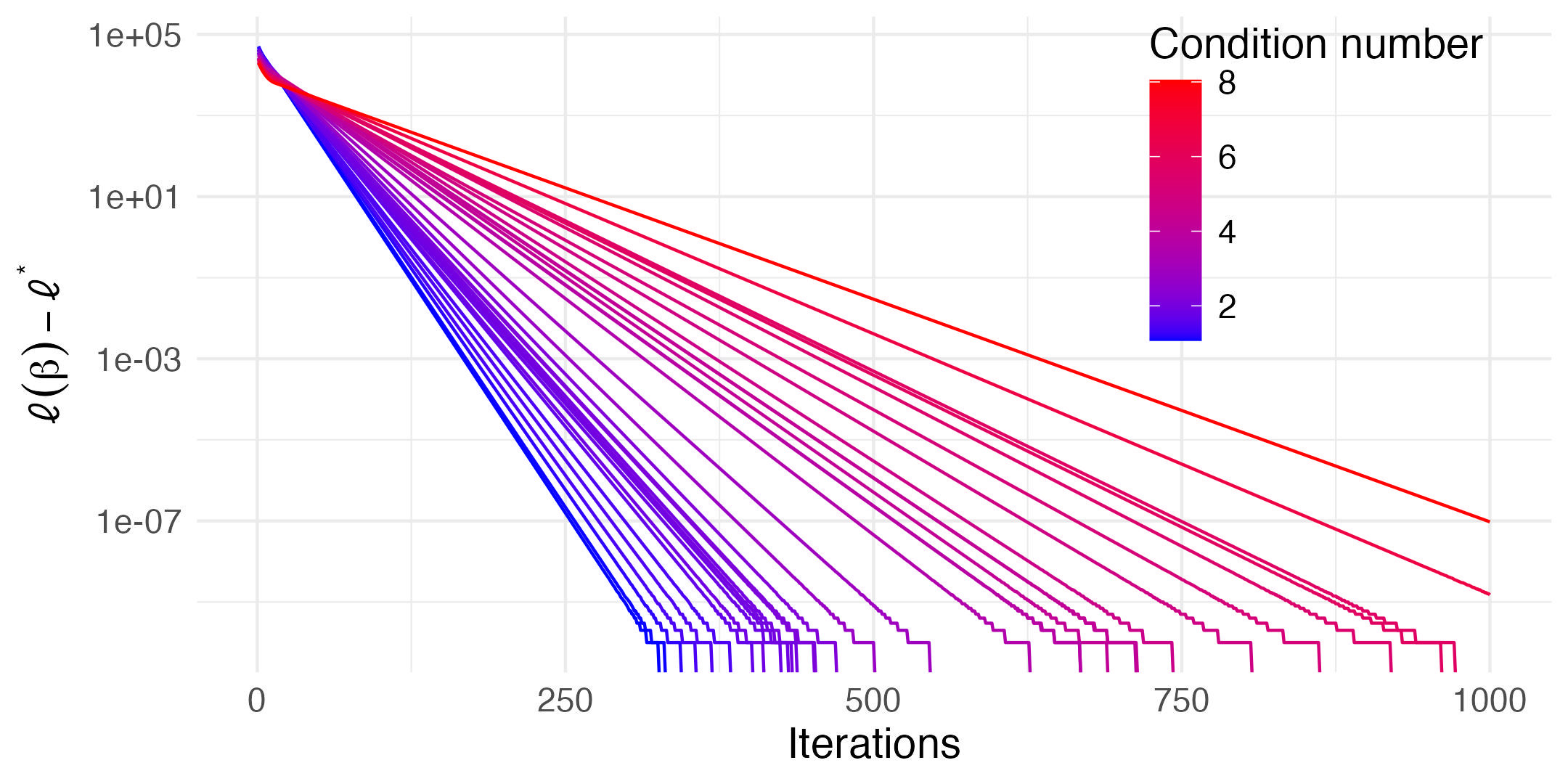}
\caption{Linear convergence for different condition numbers (indicated by the color) for a linear model with two predictor variables. The condition numbers are induced by pairwise correlation of the predictor variables. Y axis on a logarithmic scale.}\label{fig:conv_condnum}
\end{center}
\end{figure}

\subsection{Convergence of penalized boosting to unpenalized model}{\label{app:exper_conv_unpen}}

\begin{figure}[ht!]
\begin{center}
\includegraphics[width=0.6\columnwidth]{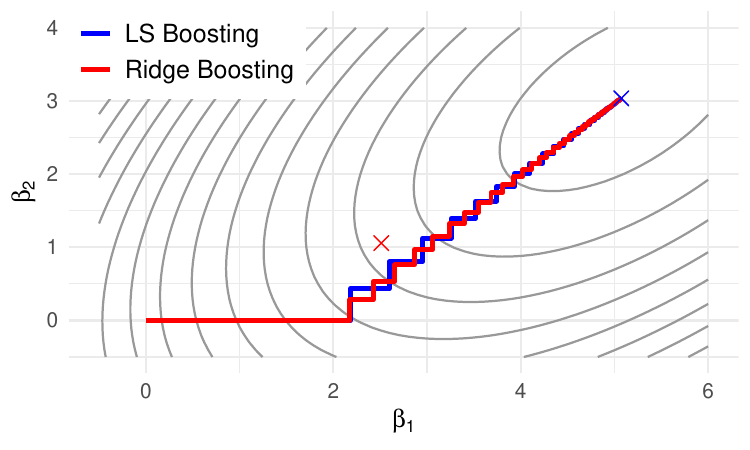}
\caption{Parameter paths for linear model (blue line) and ridge (red line) boosting as well as their actual solutions (crosses in resp.~colors). Linear model boosting is also known as Least Squares (LS) Boosting.} \label{fig:LSvsRidge}
\end{center}
\end{figure}

\paragraph{Boosting with Ridge Penalty} In our first experiment, we perform least-squares (LS) and ridge boosting on a two-dimensional problem $\beta\in\mathbb{R}^2$ to check whether both are converging against the same solution. As depicted in Figure~\ref{fig:LSvsRidge}, this is in fact the case, despite the exact ridge solution (without boosting) being far away from the LS solution.

\begin{figure}[ht!]
\centering
    \includegraphics[scale=0.5]{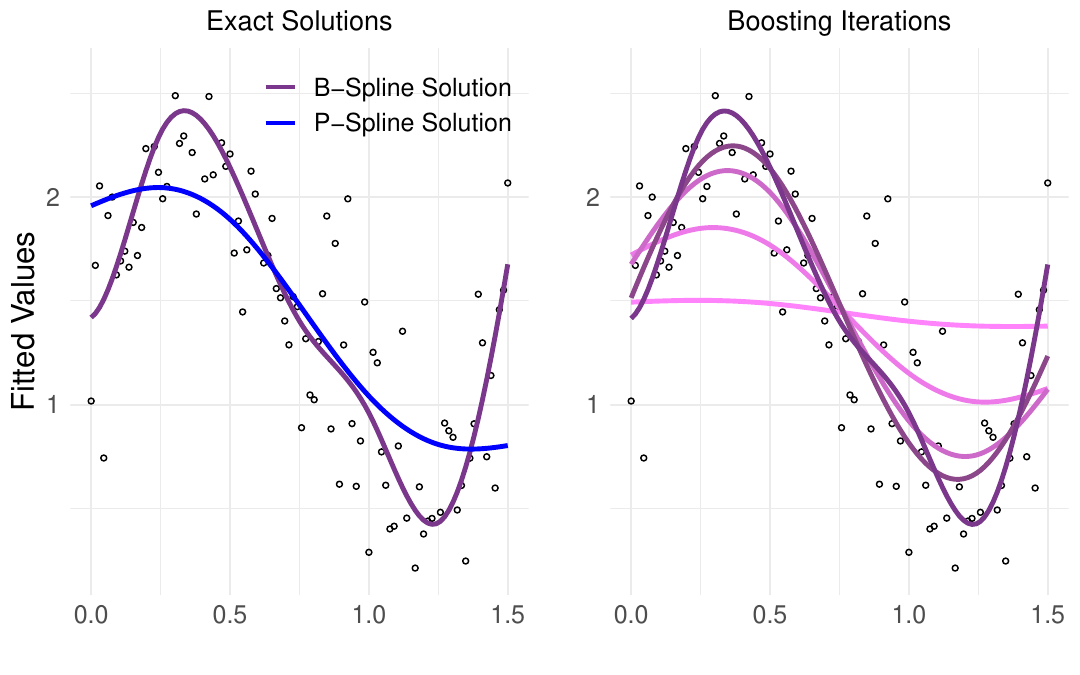}
    \vskip -0.15in
\caption{Functional Fitting via P-spline boosting on synthetic data. Left: Exact B-spline (purple) and P-spline solution (blue). Right: P-spline boosting iterates converging to the unpenalized (B-spline) solution (darkest color).}\label{fig:Psplines}
\end{figure}

\paragraph{Boosting with P-Splines} As a second example, we fit a univariate spline problem using unpenalized and penalized B-splines (B-/P-splines, resp.) using boosting. We do this both on observed data (Figure~\ref{fig:p-spline}) and on synthetic data (Figure~\ref{fig:Psplines}). The P-Spline in Figure~\ref{fig:p-spline} uses a B-Spline with 9 basis functions of degree 3 and second-order difference penalty with a penalty parameter $\lambda$ of 1. The boosted version of the P-Spline are depicted in the left plot of Figure~\ref{fig:p-spline} at boosting iterations 1, 10, 30, 60 and 5000. Similarly, The P-Spline in Figure~\ref{fig:Psplines} uses the same type of B-Spline but with a penalty parameter $\lambda$ of 10. The boosted version of the P-Spline in Figure~\ref{fig:Psplines} are shown at boosting iterations 1, 10, 30, 60 and 50000. Analogous to the result of ridge boosting, we can observe in both cases that boosting converges to the unpenalized B-spline solution despite iteratively solving a penalized LS criterion. Interestingly, due to the stage-wise fitting nature of boosting, the parameter and function path do not have to visit the penalized fit. Thus, stopping boosting iterations early does not guarantee that we can recover the penalized fit. 

%
As a consequence, the common notion of boosting performing an implicit smoothing parameter selection might be flawed and the early stopped model might belong to a completely different function class. This, in turn, can have detrimental consequences for the statistical inference, likely providing false uncertainty statements.

The examples of ridge and P-spline boosting show that as the boosting iterations increase they converge toward the unpenalized fit (the OLS and the B-spline fit, respectively). This could raise the question of why to consider penalized base learners in boosting in the first place. First, convergence paths of penalized and unpenalized base learners differ and hence both variants explore different models. This can result in different final models in case boosting is stopped early. Second, certain base learners such as P-splines might not even work without penalization, e.g., when $X^{\top}X$ is not of full rank and thus not invertible while $X^{\top}X + \lambda P$ is. 

\paragraph{Models for Spatial Data}

For the spatial data application in Figure~\ref{fig:spatial}, we fit a tensor-product spline model defined by the row-wise Kronecker product of two marginal B-spline bases, each with 20 basis functions of degree 2 and first-order difference penalty. We fix an isotropic penalty for the PLS model with both smoothing parameters set to the value 5. For the boosting model, we take the knots and basis functions as defined by the PLS model and define a corresponding base-learner. The penalty strength is defined by letting the hat matrix of both P-spline dimensions have 4 degrees-of-freedom, from which the corresponding $\lambda$ penalty values are computed. Note that these are not the effective degrees-of-freedom and the final penalization, but only define the base-learners a-priori flexibility. 

\paragraph{Boosting Function-on-Function Regression} Lastly, we demonstrate the fact that BAMs are converging to the unpenalized model fit, along the more complex application of boosting of function-on-function regression. Modeling functional relationships has recently sparked renewed interest due to its connection to in-context learning of transformers \citep[see, e.g.,][]{NEURIPS2023_e9b8a336}. 
We start by simulating a two-dimensional functional weight $\beta(s,t) = \sin(2 |s-t|) \cos(2t)$. We then create a non-linear process $x(s), s\in[0,1]$ by spanning B-spline basis functions across the domain $[0,1]$ and drawing random basis parameters from a standard Gaussian distribution. 
\begin{figure}[ht!]
\centering
\includegraphics[width=1\linewidth]{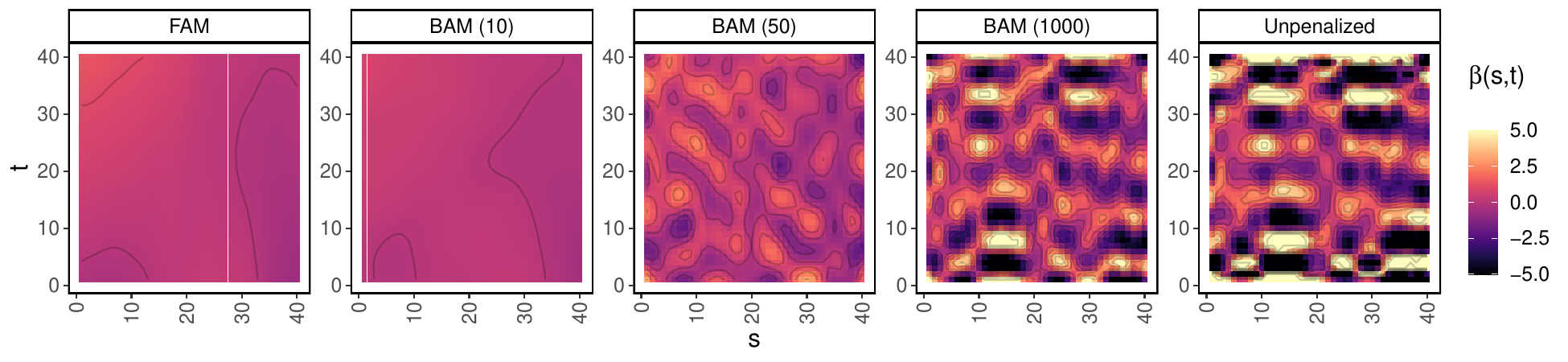}
\caption{\label{fig:fun} 
Exemplary estimated weight surfaces of a penalized functional regression (FAM; left) compared to BAM (second to fourth plot) with different numbers of iterations in brackets. For larger iterations, the estimated weight surface of BAM converges to the one of the unpenalized model (right plot).}
\end{figure}
Finally, the functional outcome is given by $y(t) = x(s)\beta(s,t) + \varepsilon(t), t\in[0,1]$, where $\varepsilon(t)$ is a white noise process. After creating $n=300$ pairs of functions, discretized to 40 time points, we fit a functional additive model \citep[FAM;][]{scheipl2015functional}, the pendant of a penalized spline regression for functional data, and compare it with BAMs for functional data \citep{fdboost}. As these models' feature matrix can be represented by a Kronecker product of evaluated basis functions and their penalization by a quadratic penalty with penalty matrix defined by a Kronecker sum, results from Section~\ref{sec:quadraticpen} apply.

Figure~\ref{fig:fun} shows the estimated weight surface $\beta(s,t)$ of FAM, boosting after 10, 50, and 1000 iterations, and the unpenalized fit. Results clearly indicate that boosting can roughly match the estimation of FAM, but when running the algorithm further will --- despite explicit penalization --- converge to the unpenalized model fit.
\paragraph{Data used for Figure \ref{fig:p-spline} and Figure \ref{fig:spatial}} In these two figures, we analyze a spatially granular data set of coronavirus disease spread from \cite{wahltinez2022covid} with a particular focus on spatial and temporal effects. The Covid-19 infection prevalence is modeled spatially over the entire US in Fig. \ref{fig:spatial}, and for San Francisco over time in Fig. \ref{fig:p-spline}. Both clearly demonstrate the convergence of penalized base learners to the unpenalized fit in accordance with \cref{prop_pspline_joint}.

\subsection{Exponential Family Boosting} \label{app:glm_exp}

\begin{figure}[ht!]
    \centering
    \includegraphics[width=0.9\textwidth]{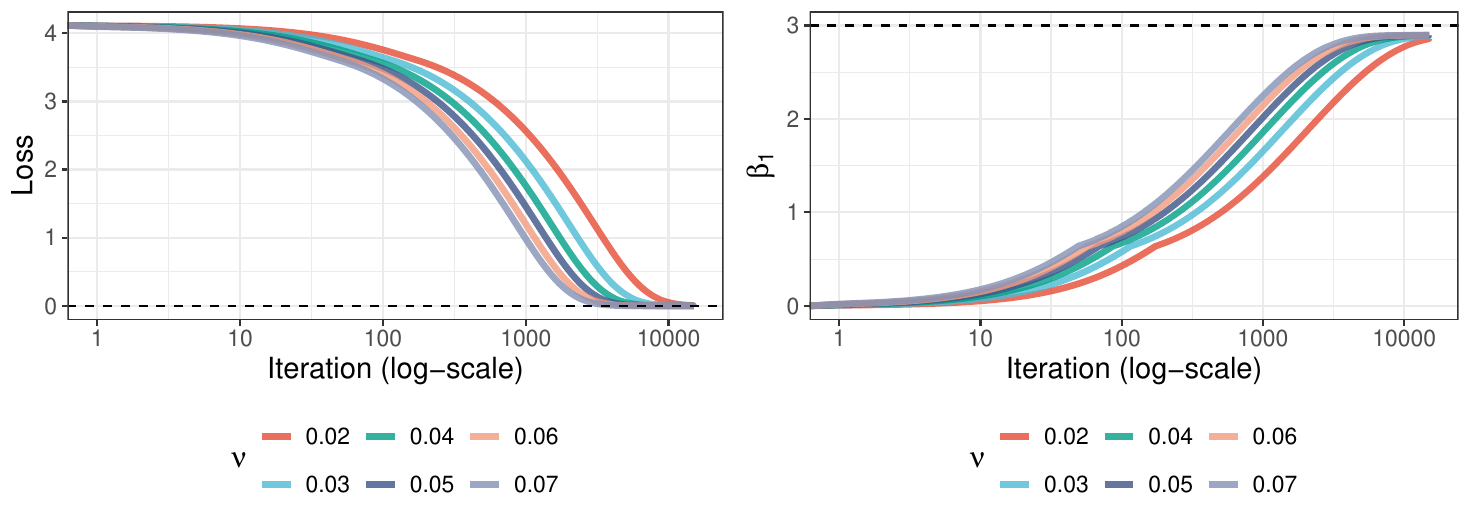}
    \caption{Binomial Boosting with different learning rates on synthetic data. Left: Loss path for different learning rates (colors) showing convergence of Binomial BAMs for different learning rates. Right: corresponding $\beta_1$ parameters.}
    \label{fig:convergence_binomial}
\end{figure}

\paragraph{Simulation Details} We simulate the data with $n=100$ and $p=2$ two features with feature effects $\beta=(3,-2)^\top$. The features are drawn from a multivariate Gaussian distribution with an empirical correlation of $\rho = 0.5$. The outcomes are finally generated by transforming the predictor $f=X\beta$ by the canonical link function $h$ ($\exp$-function in the Poisson case and the sigmoid function in the Binomial model) and drawing observations from the respective distributions with mean $h(f)$.
\paragraph{Binomial Boosting} The plot for the binomial distribution depicting the convergence on the loss and parameter level for the experiment in Section~\ref{sec:exp_glm} is given in Figure~\ref{fig:convergence_binomial}.

\paragraph{Binomial Boosting on Observational Data} We also demonstrate the convergence of Binomial Boosting on the \textit{diabetes} dataset (Pima Indians Diabetes Database) that was originally obtained from the National Institute of Diabetes and Digestive and Kidney Diseases and is available from the UCI repository \citep{blake1998uci}. We use the corrected version of the dataset that is available from the \texttt{mlbench} package \citep{mlbench2024}. The aim is to predict the existence of diabetes for patients given several diagnostic measurements, e.g., measurement of plasma glucose concentration (glucose). We standardize all predictors to have unit variance before applying the boosting procedure. For brevity, we plot only the parameter of the predictor \textit{glucose} over the boosting iterations.

\begin{figure}[ht!]
    \centering
    \includegraphics[width=0.9\textwidth]{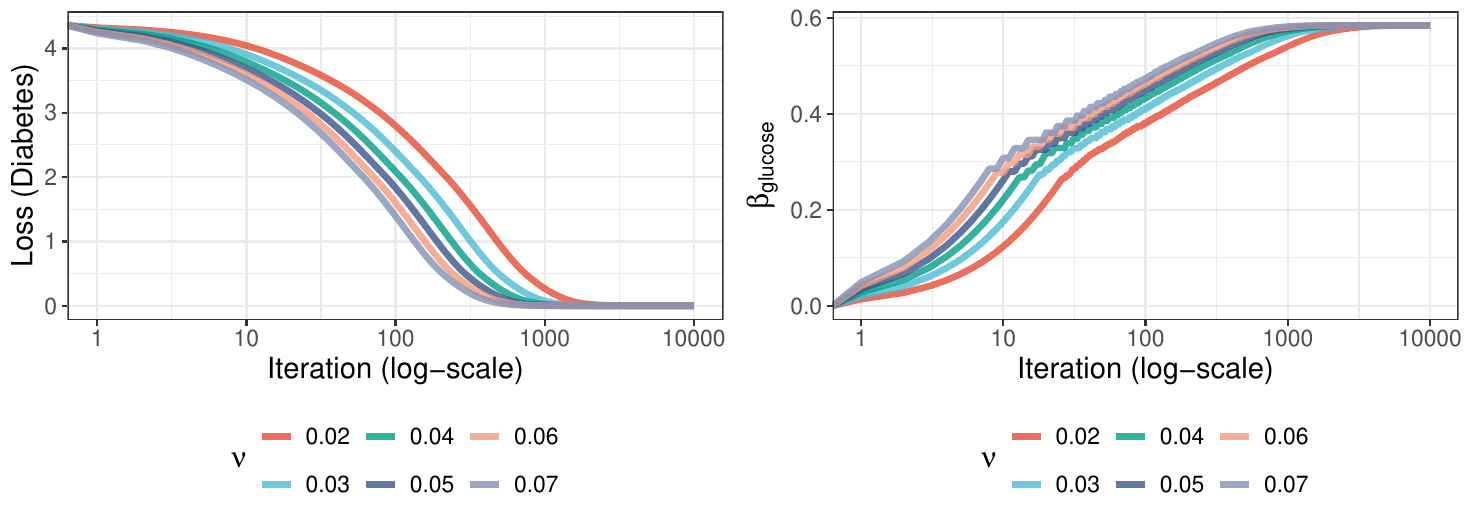}
    \caption{Binomial Boosting with different learning rates (diabetes). Left: Loss path for different learning rates (colors) showing convergence of Binomial BAMs for different learning rates. Right: Parameter of the predictor glucose.}
    \label{fig:convergence_binomial_diabetes}
\end{figure}

\begin{figure}[ht!]
    \centering
    \includegraphics[width=0.9\textwidth]{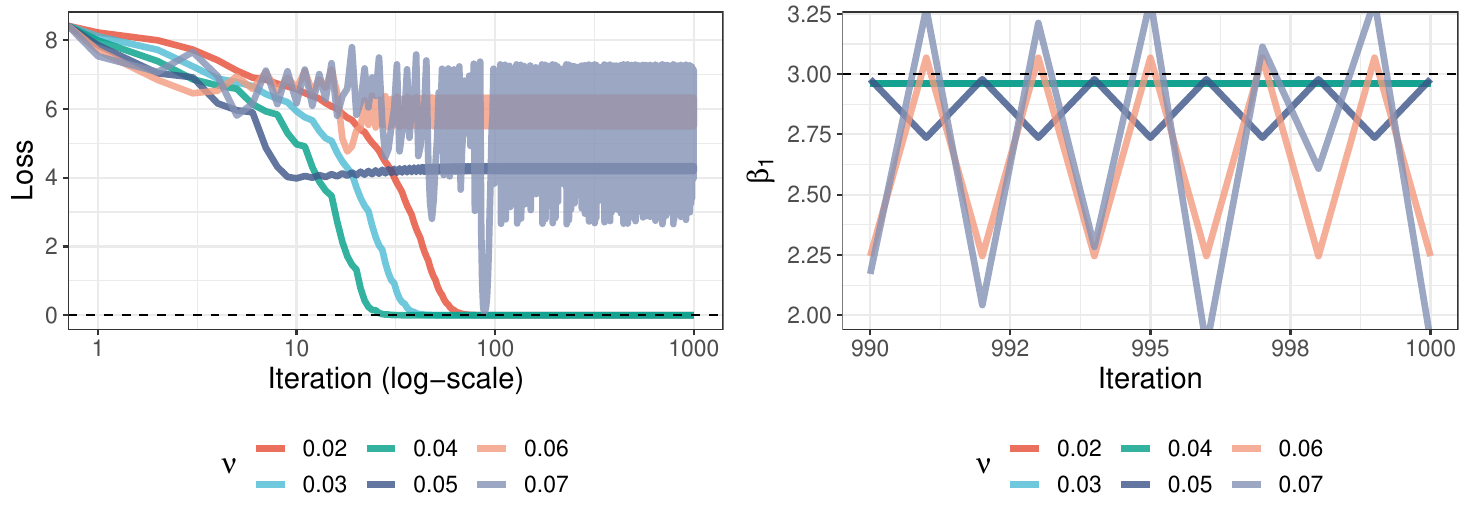}
    \caption{\textit{Poisson Boosting with different learning rates on synthetic data. Left: Loss path for different learning rates (colors) showing convergence for Poisson BAMs with the three smaller rates and non-convergence for the larger ones. Right: Last 10 updates of the corresponding $\beta_1$ parameter, depicting oscillating updates for $\nu \geq 0.05$ due to inaccurate curvature approximation.}}
    \label{fig:convergence_poisson}
\end{figure}

\begin{figure}[ht!]
\begin{center}
\includegraphics[width=0.9\textwidth]{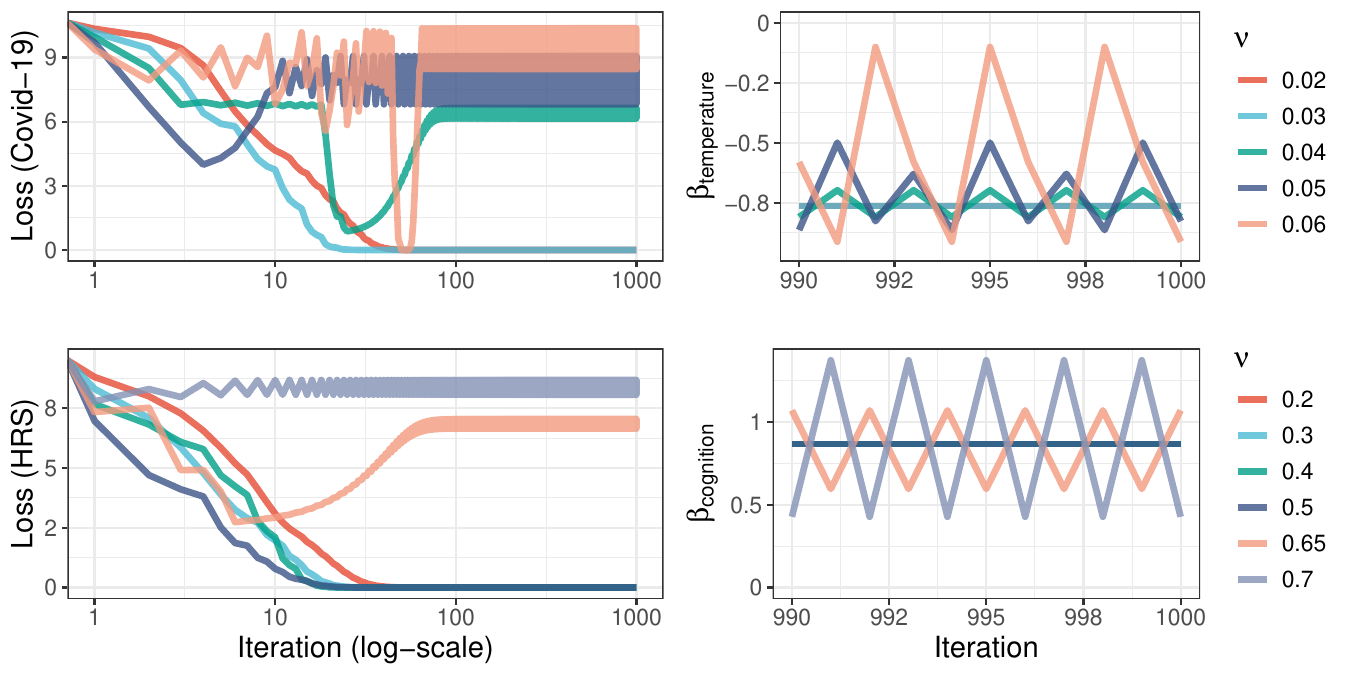}
    \vskip -0.1in
\caption{\textit{Poisson Boosting with different learning rates on observational data. Top: Poisson Boosting of Covid-19 cases in San Francisco. Bottom: Poisson Boosting of health scores in a Health and Retirement Study (HRS). Non-convergence occurs for parameters of variables temperature (top right) and cognition assessment (bottom right).}}\label{fig:convergence_poisson_obs_dat}
\end{center}
\vskip -0.1in
\end{figure}

\paragraph{Poisson Boosting} The plot for the Poisson distribution depicting the (non-)convergence on the loss and parameter level for the experiment in Section~\ref{sec:exp_glm} is given in Figure~\ref{fig:convergence_poisson}.
%
\paragraph{Poisson Boosting on Observational Data} We also demonstrate the potential non-convergence of boosting Poisson models on observational data. Boosted poisson models were used to model the number of new Covid-19 cases in San Francisco given features such as temperature (Figure \ref{fig:convergence_poisson_obs_dat}, top), and to model a person’s health score given several features such as a cognition assessment of that person (Figure \ref{fig:convergence_poisson_obs_dat}, bottom). While the former uses the same data from \cite{wahltinez2022covid} previous experiments, the latter uses the Health and Retirement Study (HRS) longitudinal data set (\url{https://hrs.isr.umich.edu}). Figure \ref{fig:convergence_poisson_obs_dat} depicts boosting’s (non)-convergence both in terms of loss and parameters for different learning rates. This is in line with the findings of Figure \ref{fig:convergence_pois_binom} in the main text.

\subsection{Cox Proportional Hazards Model Boosting} \label{app:surv_boost_exp}

\begin{figure}[ht!]
\begin{center}
\includegraphics[width=0.9\textwidth]{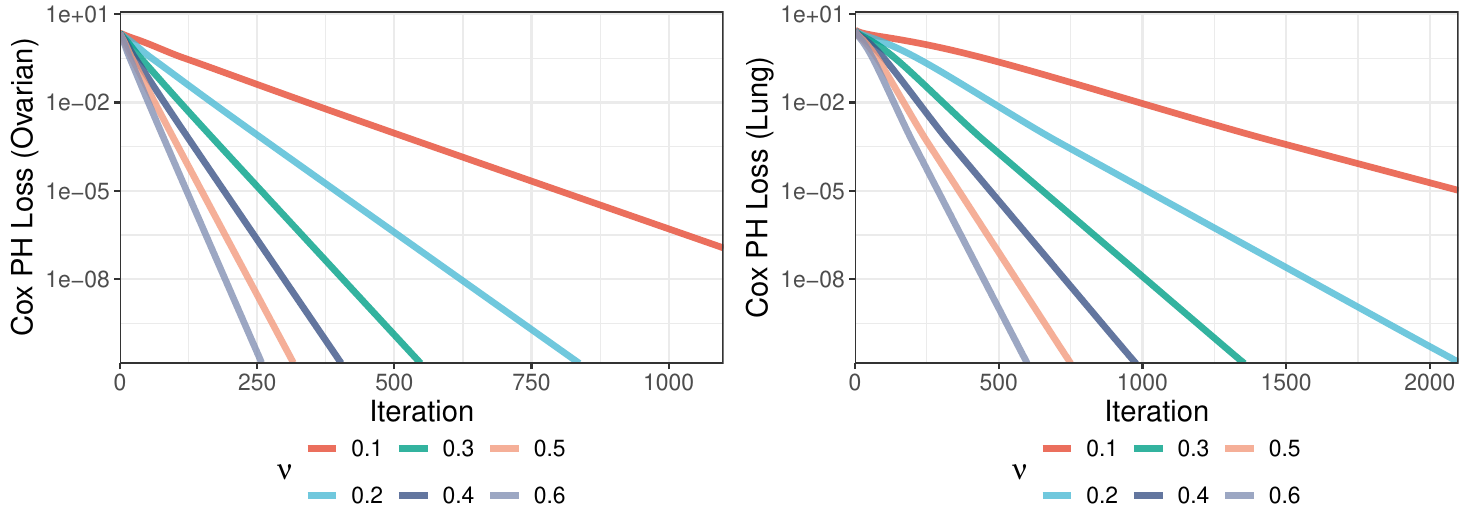}
    \vskip -0.1in
\caption{\textit{Cox PH model boosting with different learning rates (y-axis on log-scale). Left: Cox PH Boosting of survival in patients with ovarian cancer (Ovarian). Right: Cox PH boosting of survival in patients with lung cancer (Lung).}}\label{fig:conv_surv_boost}
\end{center}
\vskip -0.1in
\end{figure}

Cox Proportional Hazards (Cox PH) model boosting as mentioned in \cref{sec:glmboosting} extends the idea of gradient boosting to survival modeling. Models are again fitted using the corresponding boosting implementation of the \texttt{mboost} package. More specifically, we use boosted Cox PH models to model the survival data of patients with ovarian cancer and lung cancer, respectively. The data is provided in the \textit{Ovarian} \citep[Ovarian Cancer;][]{edmonson1979different} and \textit{Lung} \citep[North Central Cancer Treatment Group Lung Cancer;][]{loprinzi1994prospective} data sets. Figure~\ref{fig:conv_surv_boost} shows convergence with linear convergence rate (y-axis on log-scale) in the Cox PH loss\footnote{For better readability, we show the difference to the optimal loss value} over the boosting iterations.

\newpage
\subsection{Distributional Boosting} \label{app:dist_boost_exp}

\begin{figure}[ht!]
\begin{center}
\includegraphics[width=0.9\textwidth]{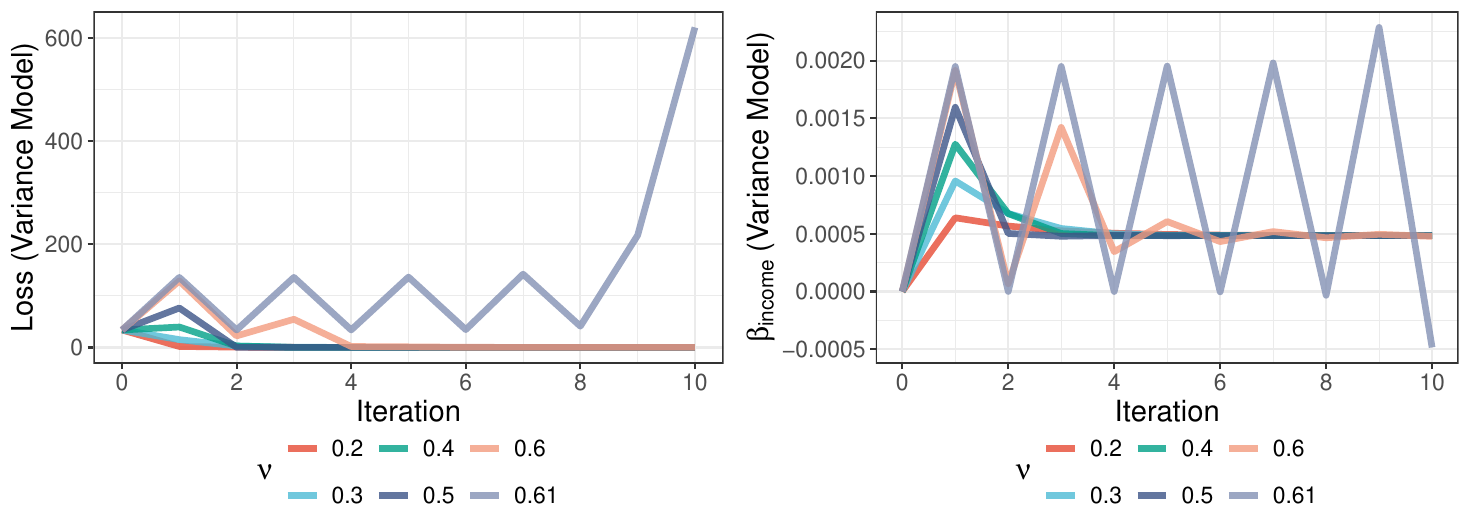}
    \vskip -0.1in
\caption{\textit{Gaussian distribution boosting with different learning rates on the Engel data set. Left: Loss w.r.t.\ the variance component (Var.\ Model). Right: Estimated parameter of income variable in the variance model. Divergence occurs already after a few boosting iterations for the larger learning rates.}}\label{fig:engel_dist_boost}
\end{center}
\vskip -0.1in
\end{figure}

\begin{figure}[ht!]
\begin{center}
\includegraphics[width=0.9\textwidth]{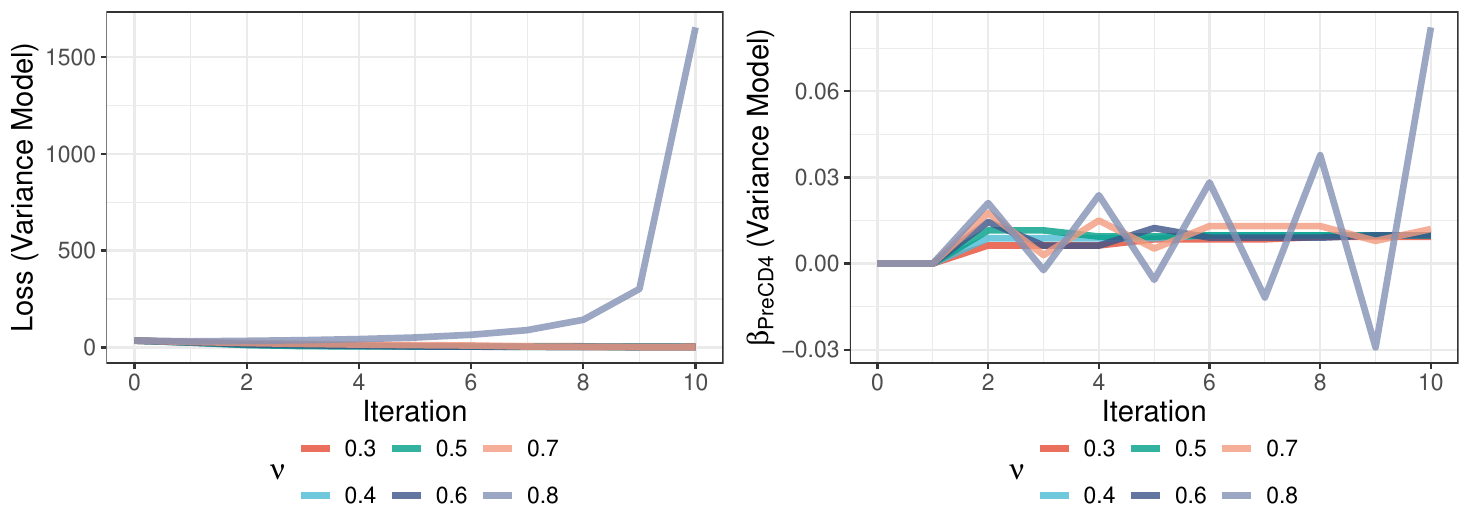}
    \vskip -0.1in
\caption{\textit{Gaussian distribution boosting with different learning rates on the CD4 data set. Left: Loss w.r.t.\ the variance component (Var.\ Model). Right: Estimated parameter of the PreCD4 variable in the variance model. Divergence occurs already after a few boosting iterations for the larger learning rates.}}\label{fig:cd4_dist_boost}
\end{center}
\vskip -0.1in
\end{figure}

Distributional boosting in the case of Gaussian distributions has been discussed in \cref{sec:dist_boost}. Figure~\ref{fig:engel_dist_boost} and Figure~ \ref{fig:cd4_dist_boost} demonstrate the potential divergence of Gaussian distribution boosting along the example of the \textit{Engel} \citep{koenker1982robust} and \textit{CD4} \citep{diciccio1996bootstrap} data set, respectively. Both data sets show a heteroscedastic variance of the dependent variable over the covariates, making them a fitting example for distributional boosting. The \textit{gamboostLSS} implementation \citep{gamboostlss} 
is used to fit boosted models for the mean and variance parameters (characterizing the Gaussian distribution of the dependent variables) for each data set. The procedure uses greedy block/component-wise updates to build the mean and variance model and alternates updates between the two models in a cyclic fashion (as described in \cref{sec:boost_add_models}). In both \cref{fig:engel_dist_boost,fig:cd4_dist_boost} the loss with respect to the variance component (variance model) diverges already after a few boosting iterations for larger learning rates. As the loss of the variance model is not globally $L$-smooth, parameter updates can become unbounded for large learning rates and thus lead to divergence. The latter is depicted in the right plots of \cref{fig:engel_dist_boost} and \cref{fig:cd4_dist_boost}.

\subsection{Remarks on the Step Size Choice} \label{app:path}

The step size that is required to guarantee convergence of BAM procedures depends both on the considered BAM class and the data at hand. In general, the chosen step size needs to fulfill the Hessian upper bound condition as stated in \cref{thm:block_conv}. While this upper bound condition can easily be verified for L2 and Binomial Boosting, this is more involved for certain other BAM classes, e.g., for Poisson boosting. In such cases, a general rule of thumb is to set the step size relatively small and subsequently check whether the fitted parameters have converged and are thus non-oscillating after a sufficient amount of iterations. While small step sizes may hurt convergence speed and thus require more boosting iterations, the BAM procedure could also be reiterated with a larger step size if parameters are still being updated (in a particular direction and are thus non-oscillating) even after many iterations with the smaller step size. 

\subsection{Path Matching} \label{app:path}

In the case of joint updates, we simulate data and compare the ridge regression and BAM path. We therefore generate $n = 100$ samples with $p = 2$ predictors. The predictors $\mathbf{X}$ are generated with an empirical correlation of $\rho=0.7$ by drawing the first predictor from a standard Gaussian distribution and defining the second predictor as a linear combination of the first and another independent normal random variable to induce correlation: ${x}_i = \rho x_1 + \sqrt{1 - \rho^2} z_i$, where $z_i \sim \mathcal{N}(0, 1)$.
The true parameter vector $\beta = (3, -2)^\top$. The response vector is then generated by $y = X\beta + \epsilon$,
where $\epsilon \sim \mathcal{N}({0}, 1)$. We then perform ridge regression and Ordinary Least Squares (OLS) regression on the simulated data. We further run BAMs as implemented in the \texttt{mboost} package with penalized linear base learners and $L_2$ loss for different values of $\lambda$ and track the parameter changes over iterations. 

We find that it is possible to find specific $\nu$ and $\lambda$ combinations such that the boosting path gets very close to the ridge regression path. This is depicted in Figure~\ref{fig:ridge_crossing} (left) for $\nu=0.1$ and maximum number of steps of 10000. However, when zooming in (right), we see that for this particular setting, the paths do not align.


%

\section{COMPUTATIONAL ENVIRONMENT}\label{sec:comp_env}

All computations were performed on a user PC with Intel(R) Core(TM) i7-8665U CPU @ 1.90GHz, 8 cores, and 16 GB RAM. Run times of each experiment do not exceed one hour. The code to reproduce the results of the experiments can be found at \url{https://github.com/rickmer-schulte/Pathologies_BAMs}.

\end{document}